\documentclass[10pt,twocolumn,letterpaper]{article}

\usepackage{cvpr}      
\usepackage{amsmath}
\usepackage{amssymb}
\usepackage{mathtools}
\usepackage{amsthm}
\usepackage{url} 
\usepackage{enumitem}
\usepackage{arydshln}
\usepackage{booktabs}  
\usepackage{colortbl}  
\usepackage{graphicx}   
\usepackage{xcolor}  
\definecolor{mygray}{gray}{0.9}  
\usepackage{algorithm}
\usepackage{listings}
\usepackage{graphicx}
\usepackage{subcaption}
\usepackage{makecell}
\usepackage{multirow}

\usepackage{booktabs}
\usepackage{adjustbox}
\usepackage{etoolbox}
\theoremstyle{plain}

\theoremstyle{definition}

\theoremstyle{remark}

\usepackage{xcolor}
\usepackage{booktabs}
\usepackage{graphicx} 
\definecolor{mygray}{gray}{0.5}
\usepackage{color}
\usepackage{colortbl}
\usepackage{array}
\usepackage[table]{xcolor}
\usepackage{pifont}
\usepackage{bbm}
\usepackage[marginal]{footmisc}

\definecolor{mygray}{gray}{0.9}
\definecolor{mygreen}{RGB}{93,174,86}
\definecolor{cvprblue}{rgb}{0.21,0.49,0.74}
\usepackage[pagebackref,breaklinks,colorlinks,allcolors=cvprblue]{hyperref}

\def\paperID{6212} 
\def\confName{CVPR}
\def\confYear{2026}

\title{Thinking in Uncertainty: Mitigating Hallucinations in MLRMs \\ with Latent Entropy-Aware Decoding}

\author{
Zhongxing Xu$^{1}$\footnotemark[1] \quad Zhonghua Wang$^{1}$\footnotemark[1] \quad Zhe Qian$^{1}$\footnotemark[1] \quad Dachuan Shi$^{2}$ \quad Feilong Tang$^{1}$ \\ Ming Hu$^{1}$  \quad Shiyan Su$^{1}$ \quad Xiaocheng Zou$^{4}$ \quad Wei Feng$^{1}$ \quad Dwarikanath Mahapatra$^{5}$ \\ Yifan Peng$^{3}$ \quad Minquan Lin$^{6}$ \quad Zongyuan Ge$^{1}$ \\ \\
\textsuperscript{\rm 1}{Monash University}\quad
\textsuperscript{\rm 2}{Georgia Tech}\quad
\textsuperscript{\rm 3}{Cornell University}\\
\textsuperscript{\rm 4}{Northeastern University}\quad
\textsuperscript{\rm 5}{Khalifa University}\quad
\textsuperscript{\rm 6}{University of Minnesota}\quad \\
{\small \texttt{\{zhongxing.xu,zongyuan.ge\}@monash.edu}}
}

\begin{document}
\maketitle
\vspace{0.5cm}

\footnotetext[1]{* Equal contribution. \textbf{\url{https://mlrm-LEAD.github.io/}}}

\begin{abstract}
\quad Recent advancements in multimodal large reasoning models (MLRMs) have significantly improved performance in visual question answering. However, we observe that transition words (e.g., because, however, and wait) are closely associated with hallucinations and tend to exhibit high-entropy states. We argue that adequate contextual reasoning information can be directly extracted from the token probability distribution. Inspired by superposed representation theory, we propose leveraging latent superposed reasoning to integrate multiple candidate semantics and maintain latent reasoning trajectories. The hypothesis is that reliance on discrete textual inputs may drive the model toward sequential explicit reasoning, underutilizing dense contextual cues during high-entropy reasoning stages. Therefore, we propose constructing rich semantic representations from the token probability distributions to enhance in-context reasoning. With this goal, we present  \textbf{L}atent \textbf{E}ntropy-\textbf{A}ware \textbf{D}ecoding (LEAD), an efficient plug-and-play decoding strategy that leverages semantic context to achieve reliable reasoning. The heart of our method lies in entropy-aware reasoning mode switching. The model employs probability-weighted continuous embeddings under high-entropy states and transitions back to discrete token embeddings as entropy decreases. Moreover, we propose a prior-guided visual anchor injection strategy that encourages the model to focus on visual information. Extensive experiments show that LEAD effectively mitigates hallucinations across various MLRMs on multiple benchmarks.
\end{abstract}
\vspace{-0.6cm}

\section{Introduction}
\label{sec:intro}
Large reasoning models~\citep{guo2025deepseek,lrm-openaio1,lrm_demystifying} enhance their complex reasoning capabilities by scaling up the computational budget during inference. This allows them to generate extended reasoning chains that incorporate causal, contrastive, and self-reflective logic before arriving at a final answer. Recently, this paradigm has been expanded to the multimodal setting. Multimodal reasoning models (MLRMs)~\citep{yang2025r1onevision,wang2025vl,yuan2025vl,deng2025openvlthinker} integrate visual understanding with linguistic reasoning by constructing explicit reasoning chains, trained via reinforcement learning with verifiable rewards. However, despite these advances and their strong multimodal reasoning capabilities, MLRMs remain highly prone to hallucinations~\citep{dong2025mirage,tian2025more,liu2025more,lu2025mitigating,chung2025mllms}.

\begin{figure}[t]
\centering
\includegraphics[width=\linewidth]{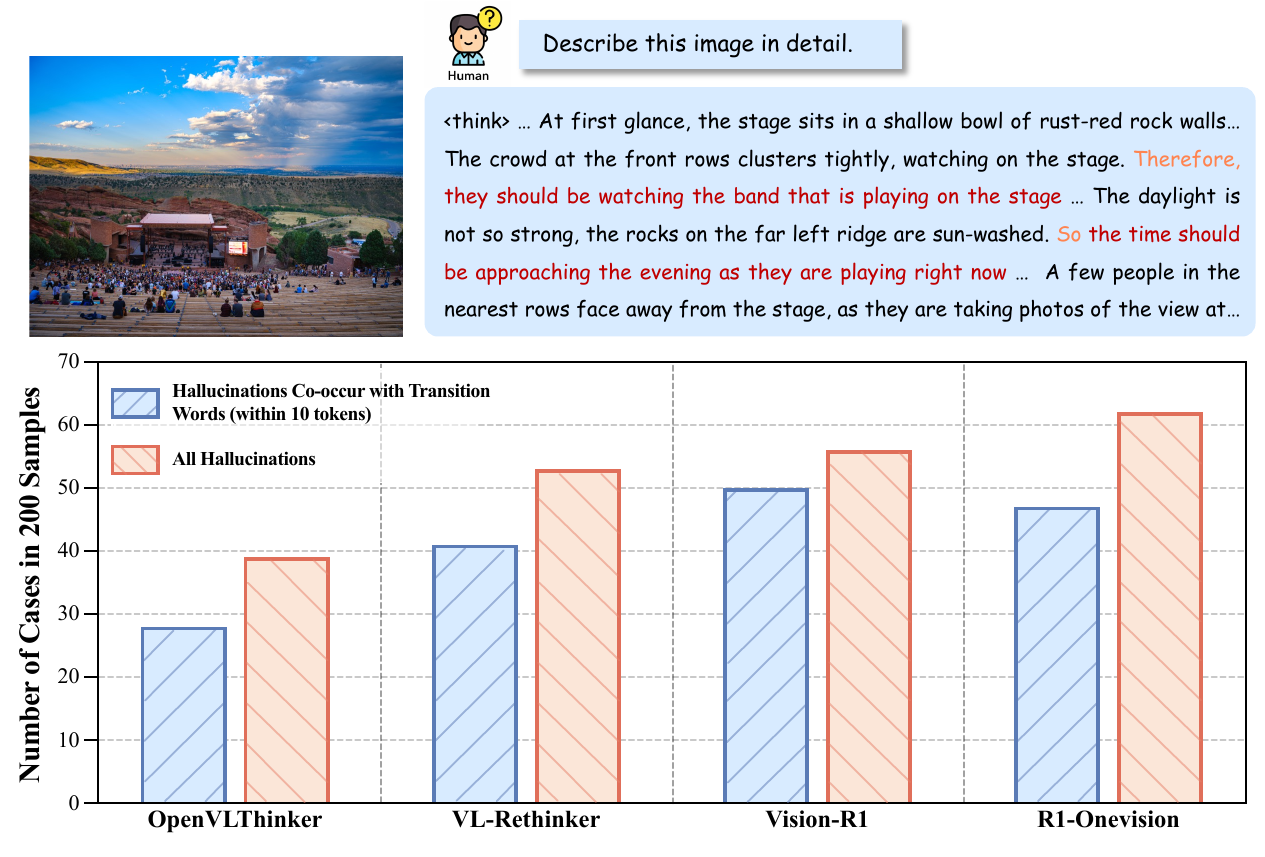}
\caption{Illustrations of the correlation between hallucinations and transition words. In MLRMs, hallucinations tend to emerge more frequently after transition words, and these cases constitute a significant proportion of the overall hallucination occurrences. }
\label{fig:figure1}
\vspace{-2em}
\end{figure}

Recent studies have primarily aimed to mitigate hallucinations in multimodal reasoning models through visual reward designs~\citep{sun2025latent,yu2025perception,ding2025vtperception} and data augmentation strategies~\citep{leng2025mmr1,chen2025advancing}, but these methods often incur substantial additional costs. Conversely, training-free decoding strategies~\citep{yin2025clearsight,huo2024self,leng2024mitigating,huang2024opera}, such as contrastive decoding, mitigate hallucinations during generation by perturbing token-level samples to adjust output distributions. Though previous works have shown effectiveness, they lack analysis of the behavioral characteristics unique to reasoning models.
In our analysis, we observe that MLRMs employ causal, contrastive, and reflective transition words (e.g., \textit{because}, \textit{however}, \textit{wait}) at significantly higher frequencies during generation. These markers help structure multimodal reasoning chains and organize semantic relations through linguistic logic, a pattern consistent with recent findings in language models ~\citep{cheng2025reasoning,wang2025beyond}. Furthermore, as shown in Fig. \ref{fig:figure1}, the content that follows such transition words often exhibits hallucinatory descriptions.

\begin{figure}
\centering
\includegraphics[width=\linewidth]{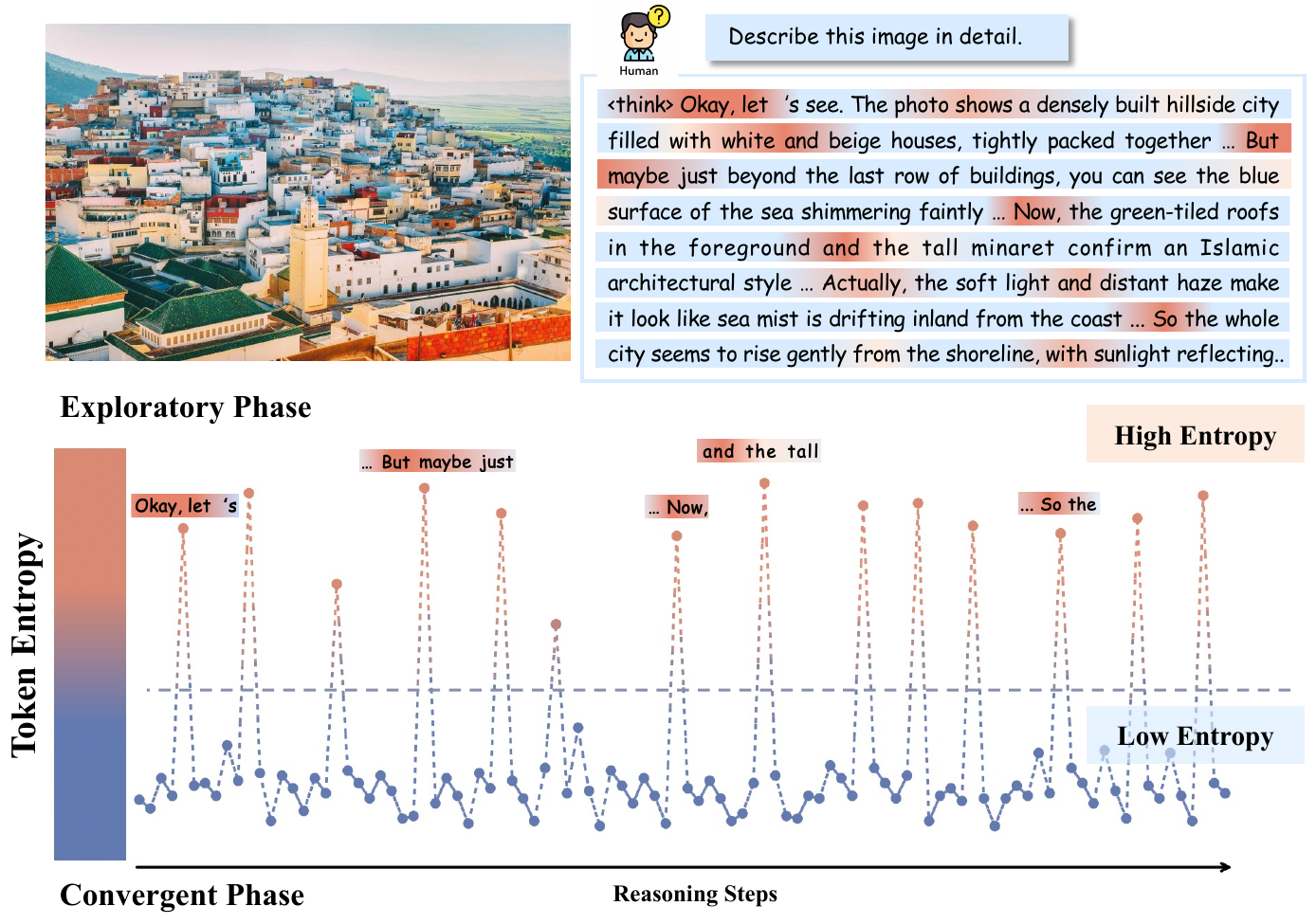}
\caption{Visualizations of token entropy during the reasoning phase show that tokens with higher entropy often correspond to transition words, consistent with our previous findings. }
\label{fig:figure2}
\vspace{-0.5cm}
\end{figure}

In this study, we investigate the intrinsic relationship between transition words and hallucinations from the perspective of token-level uncertainty, measured by entropy. As illustrated in Fig.\ref{fig:figure2}, transition words consistently exhibit higher entropy, indicating high-uncertainty stages within the reasoning chain. During these high-entropy phases, the model faces greater semantic divergence and increased competition among potential reasoning paths, thereby heightening the likelihood of hallucination. We hypothesize that reliance on discrete textual inputs encourages sequential, explicit reasoning, limiting its ability to effectively leverage dense contextual cues when uncertainty is high. In this work, we argue that the construction of richer semantic representations from token probability distributions enhances the model's contextual reasoning capability.

\begin{figure}[t]
\centering
\includegraphics[width=\linewidth]{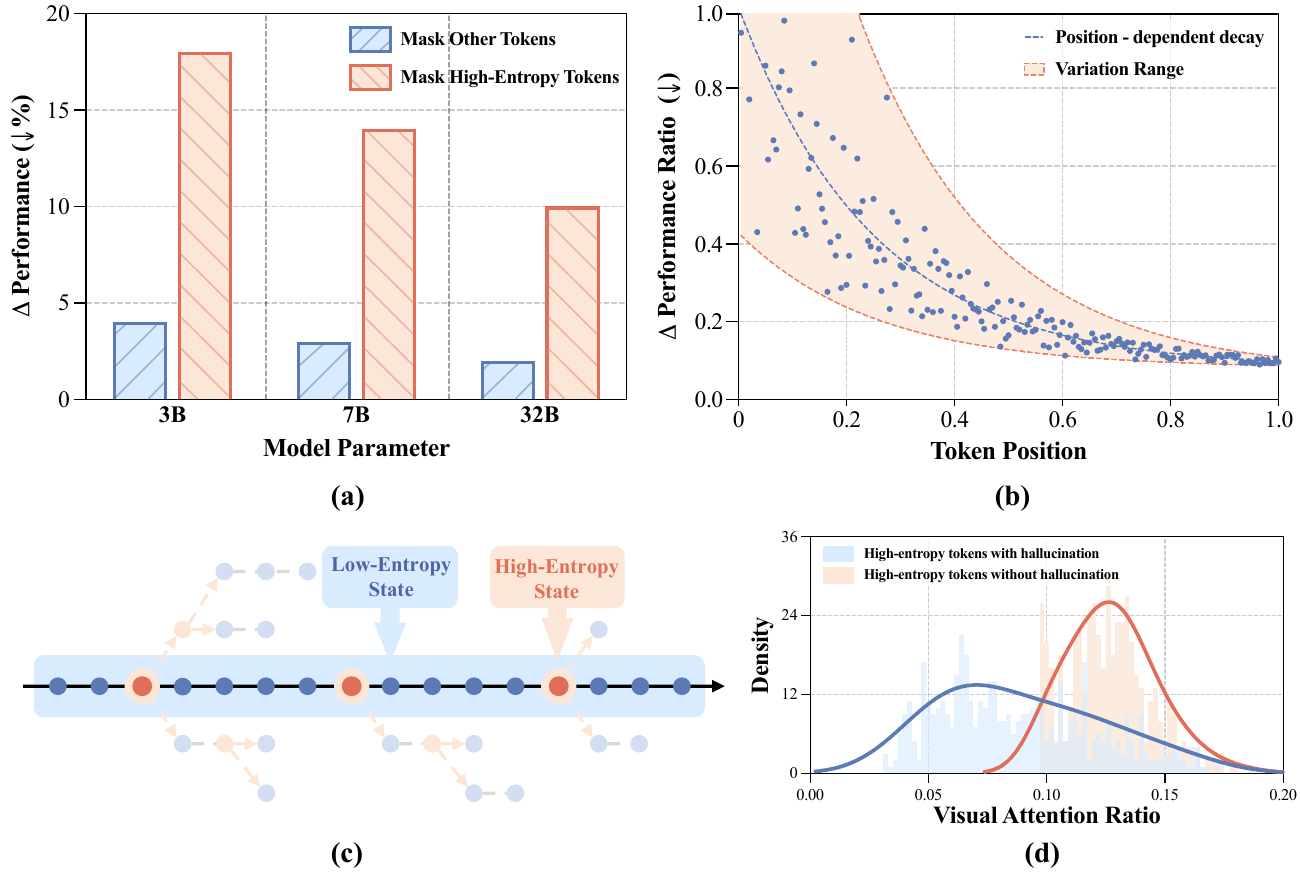}
\vspace{-0.6cm}
\caption{(a) Performance gap when masking different types of token during reasoning. Masking high-entropy tokens produces a larger performance drop than other tokens. (b) Token masking impact across reasoning steps. Earlier tokens tend to have stronger influence on the final answer, while the influence of later ones gradually diminishes. (c) Schematic depiction of reasoning paths at different states. (d) Token density comparisons. On average, high-entropy tokens without hallucinations exhibit higher visual attention ratios compared to hallucinated ones.}
\label{fig:figure3}
\vspace{-1em}
\end{figure}

To verify the role of high-entropy tokens in the reasoning chain, we conduct token masking ablation experiments. As illustrated in Fig. \ref{fig:figure3} (a), masking high-entropy tokens leads to a significant drop in reasoning performance, whereas masking low-entropy tokens causes only minor degradation. This indicates that high-entropy tokens serve as critical informational nodes in the reasoning process. We further divide the explicit reasoning chains of MLRMs into five segments and perturb high-entropy tokens in each segment. As illustrated in Fig. \ref{fig:figure3} (b), token masking applied early in the reasoning chain results in the most severe performance degradation. This finding demonstrates that early high-entropy tokens exert stronger directional influence on the overall reasoning trajectory and play a pivotal role in guiding the model toward (or away from) correct reasoning paths. Therefore, our findings suggest that maintaining semantic diversity and visual grounding during high-entropy phases is key to mitigating reasoning-related hallucinations.

In this work, we propose Latent Entropy-Aware Decoding (LEAD), a lightweight plug-and-play decoding strategy that enables reasoning reliability by leveraging contextual semantics. Specifically, when the model enters a high-entropy state, LEAD enriches the input representation by combining the discretely sampled token with its predicted probability distribution. This fuses diverse semantic cues while preserving model's inherent uncertainty. 
The core idea of LEAD is entropy-aware reasoning mode switching. Under high entropy, LEAD replaces the collapsed one-hot token vector with a probability-weighted combination of all token embeddings, implicitly preserving multiple reasoning hypotheses. As entropy decreases, the model naturally reverts to discrete token embeddings, achieving adaptive semantic convergence. 
Moreover, as illustrated in Fig.\ref{fig:figure3}(d), high-entropy tokens associated with hallucinations typically exhibit lower visual attention, suggesting a reduced reliance on visual information under high-uncertainty conditions. To address this, LEAD introduces a visual guidance vector derived from pretrained visual embeddings during high-entropy phases, encouraging the model to refocus on visual content and thus mitigating multimodal hallucinations.

With extensive experiments, LEAD demonstrates significant hallucination-mitigating performance across different MLRMs on both general and scientific multimodal reasoning benchmarks, validating its effectiveness. Our contributions are as follows: 

\begin{itemize}
    \item We analyze the relationship between transition words and hallucinations in multimodal reasoning from the perspective of token-level uncertainty.
    \item We propose LEAD, a plug-and-play decoding approach that effectively mitigates hallucinations in high-entropy reasoning states through an entropy-aware reasoning and visual injection mechanism.
    \item Extensive evaluations on both general and scientific tasks show the superior performance of LEAD, offering an effective solution for multimodal reasoning hallucinations. 
\end{itemize}

\section{Related Work}

\paragraph{Multimodal Large reasoning models.} Recent multimodal large language models (MLLMs) have achieved substantial progress in multimodal reasoning, largely driven by innovations in post-training techniques. Among these, supervised fine-tuning (SFT) ~\citep{zhang2025r1,ni2025point,wei2025advancing,weifirst,mao2025unirl,liang2025modomodo} and reinforcement learning (RL) ~\citep{tian2025more,liang2025mm,ming2025oceanr1,wang2025sota,xiao2025m2io} remain the two most common and fundamental approaches. A number of recent works ~\citep{leng2025mmr1,wang2025internvl3,zhang2025thyme,qiao2025we,tan2025reason,dong2024insight} primarily focus on enhancing long-chain reasoning in MLLMs through SFT. Meanwhile, the Group Relative Policy Optimization algorithm has emerged as a standard paradigm for training multimodal large reasoning models~\citep{liu2025visual,liu2025segzero,xiao2025fast,wang2025vl,wang2025visualprm,liuyue_GuardReasoner-VL}. Among these, some approaches~\citep{wan2025srpo,wu2025synthrl,chen2025advancing,wu2025reinforcing,hong2025glm,wei2025open,chen2025synergy,ai2025m2} adopt a two-stage training paradigm, while others directly employ reward-optimized RL strategies on large-scale datasets~\citep{wang2025vrag,wang2025vicrit,chen2025sifthinker,zhu2025shuffle,chen2025learning}.

\vspace{-1em}

\paragraph{Multimodal reasoning Hallucinations.} Despite improvements from chain-of-thought reasoning, multimodal reasoning models remain prone to hallucinations, including contradictions with visual evidence ~\citep{dong2025mirage,tian2025more,liu2025more,lu2025mitigating,chung2025mllms,li2025mixture,xu2025toward} and logical inconsistencies in reasoning ~\citep{song2025hallucination,cheng2025chain,lu2025auditing,li2025hallucination,sun2025detection,yao2025reasoning,qian2025demystifying,huang2025pear,shi2025swireasoning}. One solution is to optimize the reward-function paradigm~\citep{xiao2025advancing,yu2025perception,ding2025vtperception,wang2025perception,wei2025open} to improve perception and stabilize multimodal reasoning.
Existing multimodal hallucination mitigation methods include contrastive decoding~\cite{leng2024mitigating,wang2024mitigating,huo2024self,zhang2025self,yin2025clearsight} and self-corrective attention~\cite{huang2024opera,liu2024paying,xing2024mitigating,ma2024vista,tang2025seeing}, which reduce reliance on biases and priors. Inspired by superposed representation theory~\cite{hao2024training,zhuangmixture,zhang2025soft,wu2025llms,deng2025latent}, we propose a latent superposed reasoning approach for reasoning models, which uses the token probability distribution to extract sufficient contextual information and effectively mitigates hallucinations.

\section{Methodology}

\begin{figure*}[t]
\centering
\includegraphics[width=\linewidth]{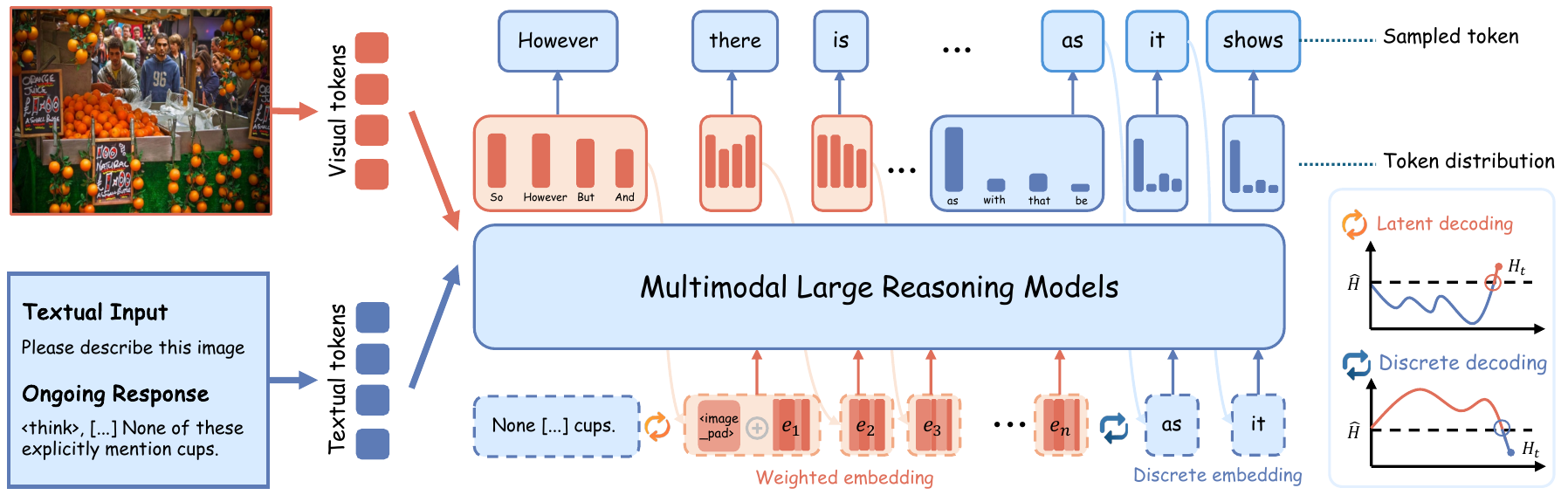}
\caption{Illustration of multimodal reasoning and entropy-aware decoding. The model receives both visual and textual tokens (left) and generates responses by integrating contextual information. During reasoning, token-level entropy $H_t$ measures model confidence and is compared with the reference entropy $\hat{H}$. High-entropy states (orange) trigger latent decoding, using probability-weighted embeddings to preserve semantic diversity, while low-entropy states (blue) activate discrete decoding, using sampled tokens for precise semantic convergence. This adaptive switching mechanism balances exploration and commitment in multimodal reasoning. }
\label{fig:figure4}
\vspace{-1em}
\end{figure*}

Figure~\ref{fig:figure4} provides an overview of the proposed strategy, which builds upon the MLRM decoding paradigm introduced in Section~\ref{sec:3.1}. Section~\ref{sec:3.2} elaborates on the entropy-aware reasoning mode switching, designed to optimize embedding representations under high-entropy states and guide the model toward semantically enriched contextual information. Meanwhile, Section~\ref{sec:3.3} introduces a guidance vector derived from the pretrained visual modality to strengthen the model's focus on visual content during uncertain reasoning phases. For clarity, Algorithm~\ref{alg:lead} exhibits the pseudocode for the decoding process of LEAD.

\subsection{MLRMs Generation} 
\label{sec:3.1}

\paragraph{Vision and Language Inputs.}
A Multimodal Large Reasoning Model (MLRM) accepts both image and text as input. The raw image is first processed by a vision encoder to extract semantic features, which are then projected into the language model's input space through a cross-modal projection module, forming a sequence of $N$ vision tokens $\mathbf{x}^{v} = \{x_{v,1}, x_{v,2}, \dots, x_{v,N}\}$. Meanwhile, the textual input is tokenized and embedded to form a sequence of $M$ text tokens $\mathbf{x}^{t} = \{x_{t,1}, x_{t,2}, \dots, x_{t,M}\}$. These vision and text tokens are concatenated to form the complete multimodal input sequence $\mathbf{x}=\mathbf{x}^v \oplus \mathbf{x}^t = \{x_t\}_{t=1}^{T}$, where $T = N + M$, serving as the input for subsequent reasoning and enabling the model to jointly process and infer over visual and linguistic information.

\paragraph{MLRMs Forward.}
The backbone of the MLRMs, denoted as $R_{\theta}$, is a pre-trained LLM parameterized by $\theta$, which generates responses autoregressively. 
Given a multimodal input $\mathbf{x}$, the model predicts the next token distribution at each time step $t$ as:

\begin{equation}
p_t = R_\theta \big(\cdot \mid  \mathbf{x}, y_{<t}\big) \in \Delta^{|\mathcal{V}|-1},
\label{eq:mlrm-forward}
\end{equation}
where $y_{<t} = (y_1,y_2,\dots,y_{t-1})$ denotes all previously generated tokens, $\mathcal{V}$ is the vocabulary of the model, and $\Delta^{|\mathcal{V}|-1}$ denotes the $(|\mathcal{V}|-1)$-dimensional probability simplex.  

\vspace{-1em}

\paragraph{Discrete Reasoning Decoding.}
Reasoning models achieve test-time scaling by explicitly separating the intermediate reasoning phase from the final answering phase. 
Given a multimodal input $\mathbf{x}$, the model first generates a reasoning trajectory $\mathbf{r}_{1:m} = (r_1, r_2, \dots, r_m)$ and then produces the final answer sequence $\mathbf{a}_{1:n} = (a_1, a_2, \dots, a_n)$, thereby structuring generation into two distinct stages.

At each intermediate reasoning step $t$, the model first computes a probability distribution $p_t$ over the vocabulary based on the multimodal input embeddings $e(\mathbf{x})$ and the embeddings of all previously generated reasoning tokens $e(r_{<t})$, and sample the token $r_t$ in current step:
\begin{equation}
p_t = R_\theta\big(e(\mathbf{x}), e(r_{<t})\big),
\quad
r_t \sim p_t, \quad r_t \in \mathcal{V}.
\end{equation}
Decoding continues until the special end-of-thinking token $\langle/think\rangle$ is generated. The model then enters the answering phase, where $\mathbf{a}_{1:n}$ is decoded in the same manner. 

\vspace{-1em}

\paragraph{Latent Reasoning Decoding.}

Although discrete reasoning improves reliability by exposing intermediate reasoning steps, its decoding strategy collapses the full predictive distribution $p_t$ into a single sampled token at each step, thereby discarding crucial distributional information that may be needed to navigate uncertain reasoning states.
To address this limitation, latent reasoning decoding replaces the discrete choice with a continuous representation that retains the entire predictive distribution. At reasoning step $t$, the model outputs a probability distribution $p_t$ over the vocabulary,
and forms a probability-weighted embedding for the next step as:
\begin{equation}
    \tilde{e}_{t} = \mathbb{E}_{v \sim p_t}\left[e(v)\right],
\end{equation}
where $\mathbb{E}$ denotes the expectation under the distribution $p_t$, and $e(v)$ denotes the embedding of token $v$. This continuous embedding, representing a mixture of all possible tokens, is fed back into the model as input for the next step, rather than the one-hot embedding of a sampled token.
Such a formulation allows the model to propagate contextual uncertainty across reasoning steps
and mitigates information loss inherent in discrete sampling.

\subsection{Entropy-Aware Reasoning Mode Switching}
\label{sec:3.2}

As shown in Figure~\ref{fig:figure3}(c), multimodal reasoning models exhibit distinct reasoning states during generation. The high-entropy phase corresponds to increased semantic uncertainty and competition among potential reasoning paths that can easily trigger hallucinations. In contrast, the low-entropy phase reflects a converging reasoning chain with more stable outputs. 
However, existing models typically operate under a fixed discrete reasoning mode and are unable to adapt to these dynamic states. To address this limitation, we propose an entropy-aware dynamic reasoning switch mechanism that uses token-level entropy as a confidence indicator. During high-entropy phases, it activates latent reasoning decoding to maintain semantic diversity; as entropy decreases, it switches back to discrete decoding to ensure stable convergence. This adaptive mechanism allows the reasoning mode to dynamically respond to uncertainty.

\vspace{-0.2cm}

\paragraph{Mode Switch Criterion.}

We use token-level entropy $H$ to measure the model's uncertainty at each generation step. 
Formally, at step $t$, the entropy is defined as:
\begin{equation}
H_t = -\sum_v p_t[v]\log p_t[v],
\label{eq:entropy}
\end{equation}
where $p_t[v]$ denotes the predicted probability of token $v$.Intuitively, high entropy arises when several candidate tokens have similar probabilities, e.g., $p_t[v_1] \approx p_t[v_2] \approx \cdots \approx p_t[v_m]$, indicating competition among multiple potential reasoning paths in the semantic space. Conversely, when a single token dominates, \ie, $p_t[v^\ast] \gg p_t[v]$ for all $v \ne v^\ast$, the model's uncertainty decreases and its reasoning process progressively converges toward a single deterministic trajectory.

Let $\hat{H}$ be the reference entropy threshold for the current reasoning mode, which is initialized at the beginning of each mode and updated after every transition. This allows the model to adjust its reasoning behavior adaptively according to the evolving uncertainty state. The model dynamically switches between reasoning modes based on the local trend of entropy variation. Specifically, the next-step input embedding $\tilde{e}_{t}$ is defined as:

\begin{equation}
\small
\tilde{e}_{t} =
\begin{cases}
e(r_t), & \text{if $H_t < \hat{H}$ (Uncertainty drops),}\\
\mathbb{E}_{v \sim p_t}[e(v)], & \text{otherwise (Uncertainty rises).}
\end{cases}
\label{eq:mode_switch}
\end{equation}
where $p_t$ is the probability distribution at current step and $r_t$ is the token sampled from $p_t$. In low-entropy states, the model employs discrete token embeddings for deterministic reasoning, while in high-entropy states, it utilizes probability-weighted embeddings to preserve semantic diversity. This entropy-aware mechanism enables a continuous, self-regulated transition between discrete and latent reasoning, with entropy serving as an internal signal.

\vspace{-1em}

\paragraph{Persistence Window.}
To avoid rapid oscillation between the two reasoning modes, we introduce a persistence window into the switching rule. Let $m_t \in \{\mathcal{D},\ \mathcal{L}\}$ denote the reasoning mode at step $t$, where $\mathcal{D}$ and $\mathcal{L}$ correspond to the discrete and latent modes, respectively. We define two gating variable for mode transition as:
\begin{equation}
    g_t^{\mathcal{D}}=\mathbbm{1}[H_t<\hat{H}],
\end{equation}
\begin{equation}
    g_t^{\mathcal{L}}=\mathbbm{1}[(H_t>\hat{H})  \land (\rho_t \ge W_{\mathcal{D}\to \mathcal{L}})],
\end{equation}
where \(\mathbbm{1}[\cdot]\) denotes the indicator function, $\rho_t$ denotes the number of consecutive steps the model has remained in its current mode, and $W_{\mathcal{D}\to \mathcal{L}}$ is the minimum number of steps the model must remain in the discrete mode before switching to the latent mode. The mode transition rule is defined as:
\begin{equation}
    m_{t+1}=g_t^{\mathcal{D}}\mathcal{D}+g_t^{\mathcal{L}}\mathcal{L}+(1-g_t^{\mathcal{D}}-g_t^{\mathcal{L}}) m_t.
\end{equation}
When a mode transition occurs, the reference entropy is updated as $\hat{H} \leftarrow H_t$, and the persistence counter $\rho_t$ is reset to 0. Otherwise, the counter is incremented as $\rho_t \leftarrow \rho_t + 1$. In practice, we enforce a persistence window only for the discrete-to-latent transition, \ie, $W_{\mathcal{D}\to \mathcal{L}}>0$. This allows a $\mathcal{L}\to\mathcal{D}$ transition to occur immediately when confidence rises. In contrast, a $\mathcal{D}\to\mathcal{L}$ transition is permitted only after the model has remained in the discrete mode for at least $W_{\mathcal{D}\to \mathcal{L}}$ steps. This asymmetric design ensures that the model stays in discrete reasoning long enough to consolidate a coherent reasoning trajectory before returning to latent exploration.

\vspace{-1em}

\paragraph{Switch Count Regulation.}

Although the model can dynamically switch between reasoning modes based on uncertainty, it may still exhibit overthinking, leading to unnecessary mode transitions even after the reasoning process has largely converged. To mitigate this, we introduce a global switch counter $\mathbf{C}_t$ with an upper bound $\mathbf{C}_{\max}$ to limit the total number of allowed mode transitions. Once this limit is exceeded, the model halts further reasoning and proceeds directly to generate the final answer.

\subsection{Entropy-Aware Visual Anchor Injection}
\label{sec:3.3}

To strengthen visual grounding during uncertain reasoning states, we introduce an entropy-aware visual anchor injection mechanism. 
Unlike continuous anchor blending, this strategy performs an injection at the first token of each high-entropy phase (\ie, at the onset of latent reasoning).  This design supplies a visual initialization cue that orients the model toward the visual semantic space without interfering with subsequent adaptive reasoning. 

Let $e_\text{vis}$ denotes the averaged embedding of pre-trained visual special tokens (\ie, \texttt{<|vision\_start|>}, \texttt{<|image\_pad|>}, \texttt{<|vision\_end|>}). When the model detects an entropy rise above the threshold $\hat{H}$ and enters the first latent step $t^\star$ in this phase, the visual anchor is injected into the weighted embedding as:
\begin{equation}
\tilde{e}_{t^\star}=(1-\lambda)\ \mathbb{E}_{v\sim p_{t^\star}}[e(v)]+\lambda\ e_\text{vis},
\end{equation}
where $\lambda \in [0,1]$ controls the strength of visual guidance. This one-time injection provides a visual grounding signal that helps stabilize the model's reasoning trajectory in the multimodal semantic space. The model injects the visual anchor each time it enters a high-entropy phase to reinforce visual guidance.

\begin{algorithm}[t]
\caption{Pseudocode of LEAD in Python Style}
\label{alg:lead}
\definecolor{codeblue}{rgb}{0.25,0.5,0.5}
\lstset{
  backgroundcolor=\color{white},
  basicstyle=\fontsize{6.0pt}{6.0pt}\ttfamily\selectfont,
  columns=fullflexible,
  breaklines=true,
  captionpos=b,
  commentstyle=\fontsize{6.0pt}{6.0pt}\color{codeblue},
  keywordstyle=\fontsize{6.0pt}{6.0pt},
}
\begin{lstlisting}[language=python]
# logits: raw scores before softmax
# E: embedding matrix
# tau: entropy threshold tracked in state
# c: maximum switch budget
# vis_injected: visual embedding injected already or not
# vis_emb/ter_emb: special embeddings via overrides

def LEAD_step(logits, E):
    # probability geometry
    p = torch.softmax(logits)
    H = -(p * (p + eps).log()).sum()

    # mode transition with threshold update
    mode = torch.where(H>=tau, LATENT, DISCRETE).where(prev)
    switched = (mode != prev)
    tau = torch.where(switched, H, tau)

    # latent embedding construction
    p = p / (p**2).sum().sqrt() + eps
    base = LATENT * (p.unsqueeze(-1) @ E).sum(dim=0) 
            + (1 - LATENT) * E[argmax_token(p)]

    # visual injection on latent embedding
    inject = base + vis_injected * vis_emb.unsqueeze(-1)

    # last embedding based on termination condition
    last_embedding = K(switch_count, c, ter_emb, inject)

    return last_embedding
\end{lstlisting}
\end{algorithm}


\section{Experiments}

\subsection{Experimental Setup}

\paragraph{Baselines.} 

We evaluate LEAD on a set of representative MLRMs, including R1-Onevision-7B~\citep{yang2025r1onevision}, Vision-R1-7B~\citep{visionr1}, VL-Rethinker-7B~\citep{wang2025vl}, VL-Cogito-7B~\citep{yuan2025vl}, and OpenVLThinker-7B~\citep{deng2025openvlthinker}. Additional results for different model scales are provided in Appendix~A.

\vspace{-1em}

\paragraph{Evaluation Benchmarks.}

We conduct evaluations on both general and domain-specific multimodal reasoning benchmarks. For general evaluation, we consider two categories:
(1) General Reasoning \& Understanding (MMEval-Pro~\citep{huang2024mmevalpro}, MMVP~\citep{tong2024eyes}, RealWorldQA~\citep{grok2beta2024}, VMCBench~\citep{zhang2025automated}, and VStar~\citep{wu2024v}) and (2) Hallucination Assessment (Bingo~\citep{cui2023holistic}, MMHalu~\citep{sun2023aligning}, and POPE~\citep{li2023pope}).
For domain-specific evaluation, we assess performance on (1) Mathematical Reasoning (MathVision~\citep{wang2024mathvision}, MathVista~\citep{mathvista}, MathVerse~\citep{zhang2024mathverse}, VisuLogic~\citep{xu2025visulogic}, Geometry3K~\citep{lu2021inter} and Mathematics subset of MMK12~\citep{meng2025mm}) and (2) Scientific Reasoning (Physics, Chemistry and Biology subsets of MMK12).

\vspace{-1em}

\paragraph{Implementation Details.}

LEAD samples tokens in the output stage using the conventional discrete manner, with the examples illustrated using the greedy decoding strategy. Details of other methods are provided in Appendix~B. For the Switch Count, we set the switching number $C_{\text{t}}$ with a default maximum value of 5. Extensive experiments indicate that $C_{\text{max}} = 5$ ensures stable and consistent generation.

\subsection{Ablation Study}

\paragraph{Effect of Entropy Threshold.}

\begin{figure}
    \centering
    \includegraphics[width=\linewidth]{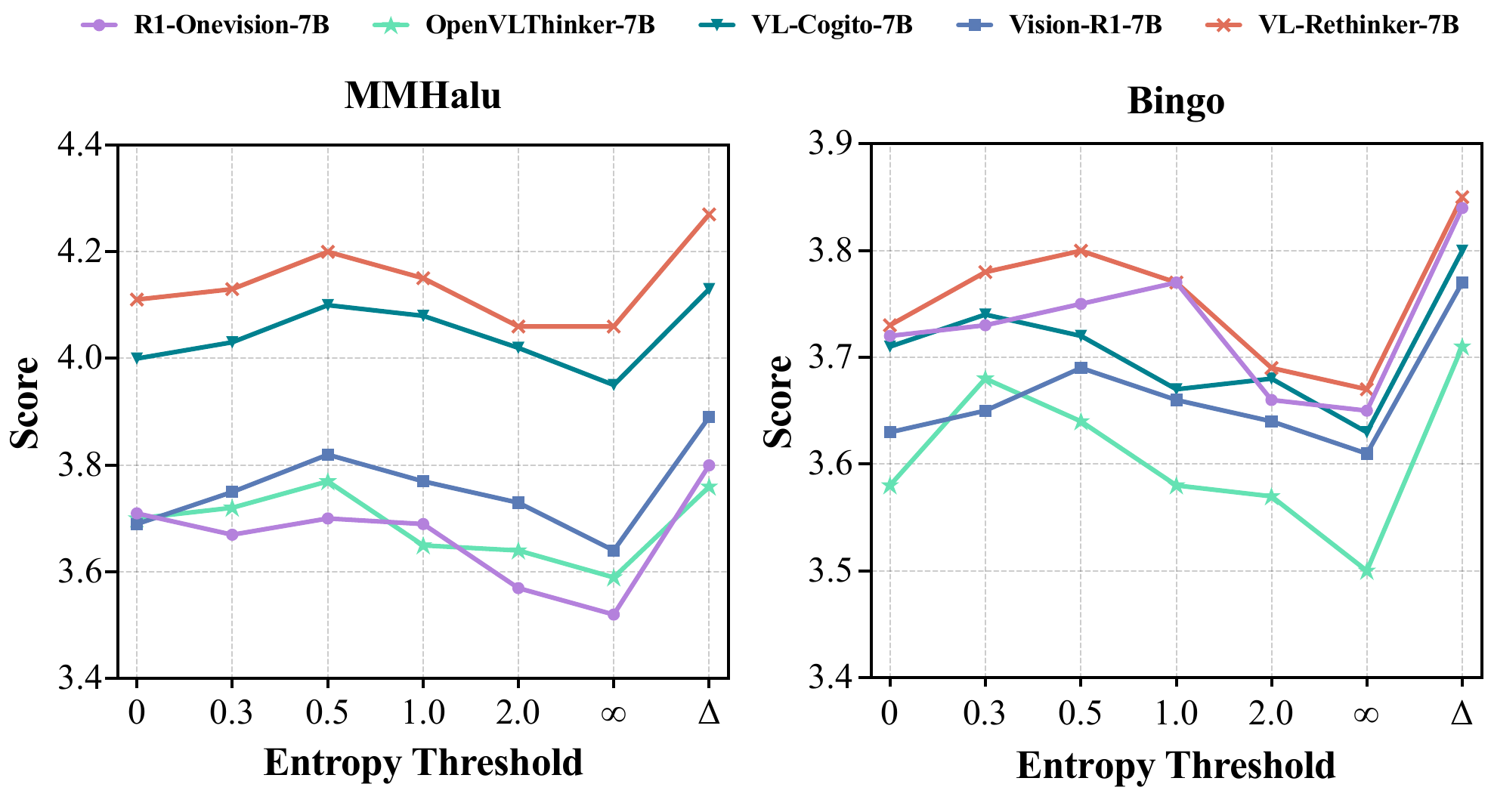}
    \vspace{-0.6cm}
    \caption{Comparisons of average score on MMHalu and Bingo datasets under different entropy thresholds. $\Delta$ denotes the dynamic thresholding strategy in LEAD. $\infty$ keeps the model in standard discrete CoT reasoning, while 0 keeps it in latent reasoning. }
    \label{fig:ablation_mixed_thinking}
    \vspace{-0.2cm}
\end{figure}

We experiment with different entropy thresholds to evaluate the effectiveness of the discrete–latent reasoning switching mechanism. 
As shown in Fig.\ref{fig:ablation_mixed_thinking}, dynamic thresholding consistently yields the best performance, improving MMHalu scores by +4.7\% and +4.1\% for R1-Onevision and Vision-R1, respectively, showing the advantage of LEAD's adaptive switching strategy. In contrast, a large threshold forces the model to remain in discrete CoT reasoning, preventing it from leveraging exploratory latent reasoning. Conversely, a small threshold keeps the model in latent reasoning for too long, weakening the discrete convergence and increasing the risk of hallucination. 

\begin{table}[t!]
\centering
\caption{Effect of visual anchor injection strength $\lambda$ on overall performance. Scores are reported for MMHalu (ranging from 0 to 6) and Bingo (ranging from 1 to 5), while accuracy is reported for VStar and MMEval-Pro. Best results are highlighted in \textbf{Bold}.}
\vspace{-0.2cm}
\label{tab:visual_injection}
\resizebox{0.48\textwidth}{!}{
\begin{tabular}{l|c|cccc}
\midrule
\rowcolor[gray]{0.9}
\textbf{Model} & $\boldsymbol{\lambda}$ & \textbf{VStar} & \textbf{MMEval-Pro} & \textbf{MMHalu} & \textbf{Bingo} \\
\midrule
\multirow{4}{*}{\textbf{R1-Onevision-7B}}
& 0   & 67.5 & 71.9 & 3.59 & 3.74 \\
& 0.2 & 69.6 & 72.0 & 3.66 & 3.73 \\
& \cellcolor{gray!20}\textbf{0.4}
  & \cellcolor{gray!20}\textbf{71.2}
  & \cellcolor{gray!20}\textbf{73.9}
  & \cellcolor{gray!20}\textbf{3.80}
  & \cellcolor{gray!20}\textbf{3.84} \\
& 0.6 & 68.1 & 73.3 & 3.77 & 3.76 \\
\midrule
\multirow{4}{*}{\textbf{Vision-R1-7B}}
& 0   & 79.1 & 72.7 & 3.69 & 3.68 \\
& 0.2 & 80.1 & 73.9 & 3.78 & 3.70 \\
& \cellcolor{gray!20}\textbf{0.4}
  & \cellcolor{gray!20}\textbf{81.7}
  & \cellcolor{gray!20}\textbf{75.1}
  & \cellcolor{gray!20}\textbf{3.89}
  & \cellcolor{gray!20}\textbf{3.77} \\
& 0.6 & 79.6 & 74.5 & 3.83 & 3.75 \\
\midrule
\end{tabular}
}
\vspace{-1em}
\end{table}

\begin{figure}
    \centering
    \includegraphics[width=\linewidth]{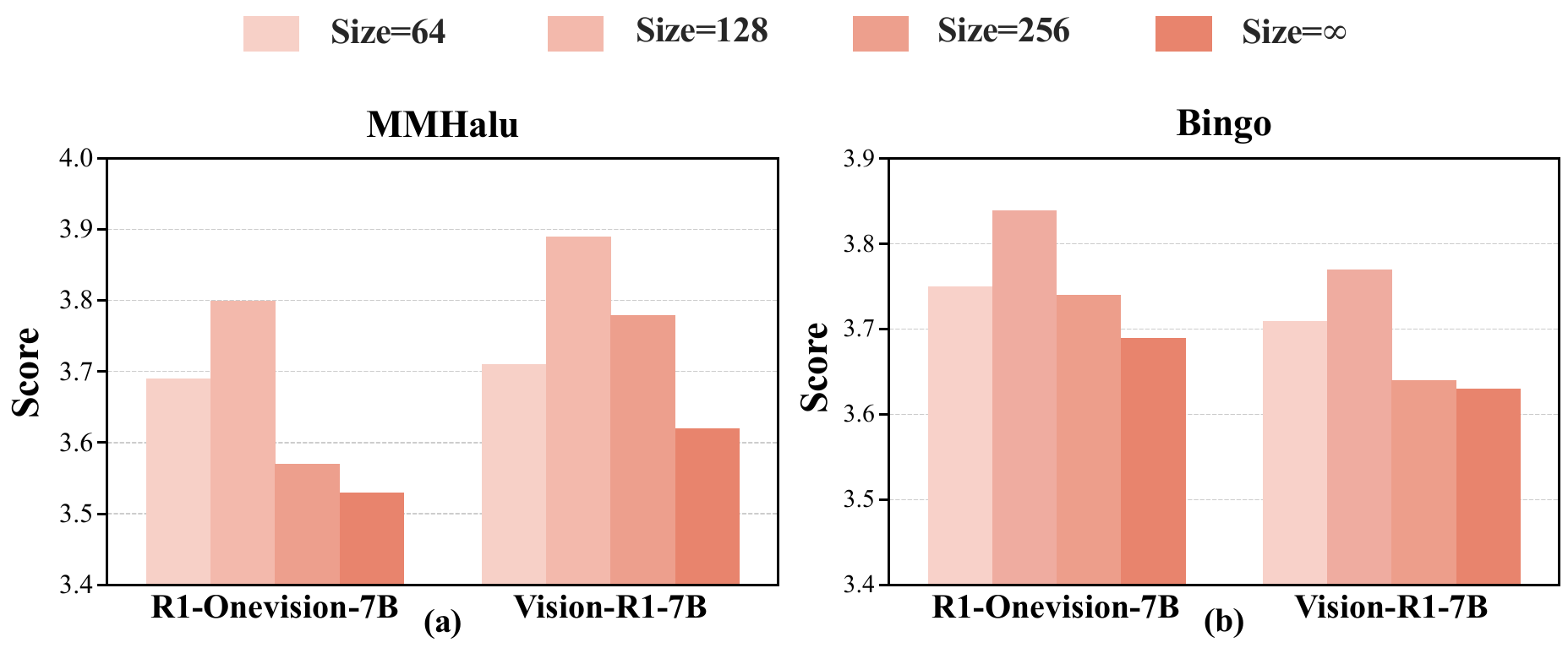}
    \vspace{-0.6cm}
    \caption{ Comparisons of model performance under different persistence window sizes. (a) and (b) show model performance with varying window values on the MMHalu and Bingo datasets.}
    \label{fig:ablation_window_size}
\vspace{-0.4cm}
\end{figure}

\begin{figure*}[t]
\centering
\includegraphics[width=\linewidth]{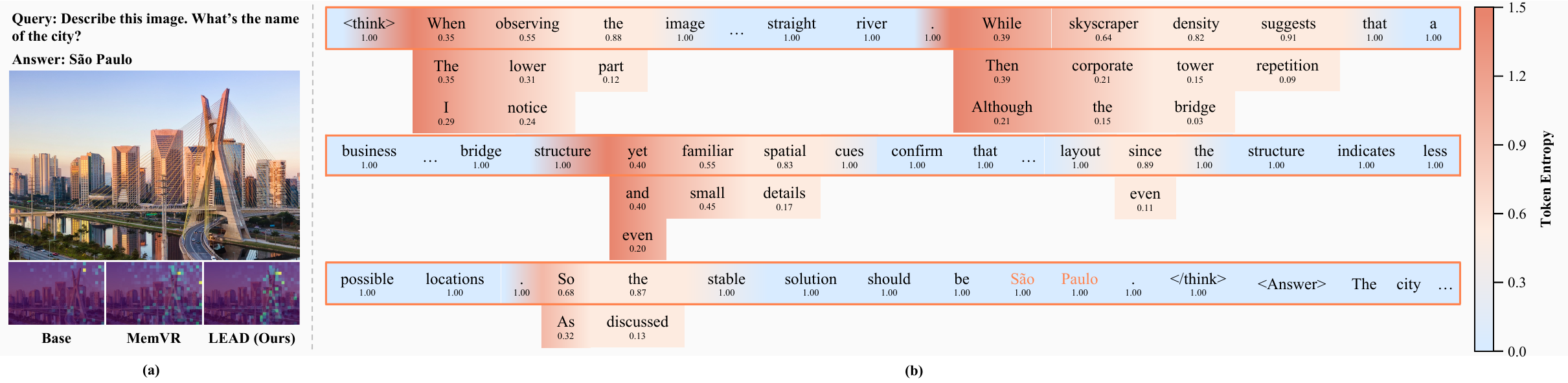}
\vspace{-0.6cm}
\caption{Qualitative visualization of LEAD under discrete and latent reasoning. (a) Comparisons of the average visual attention allocation across reasoning steps among Base, MemVR and our LEAD. (b) Example visualization of LEAD’s token-level probability distribution and entropy across reasoning steps. The token probabilities and corresponding entropies are shown at each step. The tokens highlighted in orange box correspond to those sampled in the final output sequence. More detailed visualizations are provided in Appendix D. }
\label{fig:visualization}
\vspace{-1em}
\end{figure*}

\vspace{-1em}

\paragraph{Effect of Switching Window Size.}

We examine the influence of the discrete reasoning window size on final performance. Fig.\ref{fig:ablation_window_size} shows that performance improves as the window size grows up to 128, after which it begins to decline. A moderate window size encourages the model to remain briefly in discrete reasoning before switching, thereby avoiding excessively frequent transitions. However, when the window size is too large, the model remains in discrete CoT-style reasoning for most of the inference process, reducing the benefits of latent reasoning. In the extreme case where the window size is set to $\infty$, the model switches back to discrete reasoning after its first latent reasoning turn and then remains in discrete mode permanently, causing performance to regress toward the level of standard CoT. 

\vspace{-1em}

\paragraph{Effect of Visual Anchor Injections.}

We evaluate the impact of visual anchor injection strength on hallucination mitigation. Table \ref{tab:visual_injection} presents performance across different injection strengths. Performance improves as injection strength increases, reaching its peak at 0.4 across all datasets. By injecting a moderate amount of visual information during high-entropy reasoning steps, the model is encouraged to ground its latent reasoning process in visual evidence, helping maintain consistency between generated content and the underlying image. However, when the injection strength is too high, visual embedding begins to dominate the representation, diminishing the influence of linguistic context and leading to a slight performance drop.

\vspace{-1em}

\paragraph{Qualitative Analysis.}

We visualize the response of R1-Onevision across different methods. Fig.~\ref{fig:visualization} (a) shows that LEAD allocates relatively higher visual attention to query-relevant regions compared to Baseline and MemVR. This aligns with the injection of visual anchors, which reallocates the attention to task-related visual information and reduces attention to irrelevant tokens. Fig.~\ref{fig:visualization} (b) presents the token probability distribution and token-level entropy across reasoning steps for LEAD. For clarity, we highlight the top three tokens. During latent reasoning, the token distribution appears to be more dispersed, corresponding to higher token entropy. In contrast, during discrete reasoning, the token distribution approaches a one-hot pattern with lower entropy, indicating deterministic reasoning. 

\subsection{Comparisons to State-of-the-Arts}

\paragraph{Benchmark Evaluation.}
To evaluate the general image understanding, we compare models with the LEAD extension against several decoding methods, including VCD~\citep{leng2024vcd}, MemVR~\citep{zou2024look}, and SID~\citep{huo2024self}, as shown in Table~\ref{tab:general_results}. Integrating LEAD as a plugin into R1-onevision results in an average improvement of +3.6\% in the General reasoning and understanding tasks. It also achieves significant gains in hallucination metrics, with MMHalu and Bingo scores and increasing by +4.7\% and +3.8\%, respectively. These results indicate that LEAD is effective at reducing hallucinations in unstructured environments. As shown in Table~\ref{tab:specific_results}, in domain-specific reasoning tasks, LEAD improves average accuracy by +2.0\% on mathematics benchmarks and +3.2\% on scientific benchmarks, demonstrating its effectiveness in structured and symbolic reasoning scenarios. Furthermore, the benefits of LEAD extend beyond the R1-Onevision model, as other models also experience considerable enhancements.

\begin{table*}[h]
    \centering
    \footnotesize
    \caption{Comparisons of different MLRMs with LEAD across general reasoning and hallucination benchmarks. Scores are reported for MMHalu (ranging from 0 to 6) and Bingo (ranging from 1 to 5), while accuracy is reported for all other benchmarks. }
    \vspace{-0.2cm}
    \label{tab:general_results}
    \resizebox{\textwidth}{!}{
    \setlength{\tabcolsep}{2pt}
    \renewcommand{\arraystretch}{1.3}
    \begin{tabular}{l|lllll|lllll}
        \hline
        \rowcolor{mygray}
        \multicolumn{1}{l|}{\cellcolor{mygray}} &
        \multicolumn{5}{c|}{\cellcolor{mygray}\textbf{General Reasoning \& Understanding}} &
        \multicolumn{5}{c}{\cellcolor{mygray}\textbf{Hallucination Benchmark}} \\
        \cline{2-11}
        \rowcolor{mygray}
        \multicolumn{1}{l|}{\multirow{-2}{*}{\cellcolor{mygray}\textbf{Method}}} &
        \textbf{VStar}~$\uparrow$ & \textbf{RealWorldQA}~$\uparrow$ & \textbf{MMVP}~$\uparrow$ & \textbf{MMEval-Pro}~$\uparrow$ &
        \textbf{VMCBench}~$\uparrow$ &
        \textbf{MMHalu}~$\uparrow$ & \textbf{Bingo}~$\uparrow$ &
        \textbf{POPE-R}~$\uparrow$ & \textbf{POPE-P}~$\uparrow$ & \textbf{POPE-A}~$\uparrow$ \\
        \hline

        R1-Onevision-7B & 66.5 & 62.5 & 43.0 & 69.4 & 65.2 & 3.52 & 3.65 & 84.6 & 84.0 & 82.5 \\
        + VCD           & 67.1 & 62.6 & 42.9 & 69.8 & 66.0 & 3.55 & 3.61 & 84.4 & 83.8 & 82.3 \\
        + MemVR         & 69.6 & 64.3 & 44.5 & 71.3 & 67.5 & 3.69 & 3.68 & 82.3 & 85.0 & 83.5 \\
        + SID           & 70.2 & 65.2 & 43.2 & 71.0 & 67.8 & 3.70 & 3.65 & 85.0 & 84.7 & 81.9 \\
        \rowcolor{gray!20}
        \textbf{+ LEAD (Ours)} & 71.2~\color{blue!60}{(+4.7)} & 66.4~\color{blue!60}{(+3.9)} & 45.0~\color{blue!60}{(+2.0)} & 73.9~\color{blue!60}{(+4.5)} & 67.9~\color{blue!60}{(+2.7)} & 3.80~\color{blue!60}{(+4.7)} & 3.84~\color{blue!60}{(+3.8)} & 85.9~\color{blue!60}{(+1.3)} & 85.3~\color{blue!60}{(+1.3)} & 83.9~\color{blue!60}{(+1.4)} \\
        \cdashline{1-11}
        
        Vision-R1-7B    & 78.5 & 64.3 & 44.0 & 72.2 & 80.3 & 3.64 & 3.61 & 88.0 & 85.2 & 84.0 \\
        \rowcolor{gray!20}
        \textbf{+ LEAD (Ours)} & 81.7~\color{blue!60}{(+3.2)} & 67.5~\color{blue!60}{(+3.2)} & 46.3~\color{blue!60}{(+2.3)} & 75.1~\color{blue!60}{(+2.9)} & 82.1~\color{blue!60}{(+1.8)} & 3.89~\color{blue!60}{(+4.1)} & 3.77~\color{blue!60}{(+3.2)} & 91.4~\color{blue!60}{(+3.4)} & 88.3~\color{blue!60}{(+3.1)} & 87.7~\color{blue!60}{(+3.7)} \\
        \cdashline{1-11}
        
        VL-Rethinker-7B & 67.6 & 69.3 & 42.0 & 73.2 & 73.9 & 4.06 & 3.67 & 85.5 & 81.8 & 82.8 \\
        \rowcolor{gray!20}
        \textbf{+ LEAD (Ours)} & 70.1~\color{blue!60}{(+2.5)} & 71.2~\color{blue!60}{(+1.9)} & 46.6~\color{blue!60}{(+4.6)} & 75.7~\color{blue!60}{(+2.5)} & 75.2~\color{blue!60}{(+1.3)} & 4.27~\color{blue!60}{(+3.5)} & 3.85~\color{blue!60}{(+3.6)} & 86.2~\color{blue!60}{(+0.7)} & 85.1~\color{blue!60}{(+3.3)} & 84.9~\color{blue!60}{(+2.1)} \\
        \cdashline{1-11}
        
        VL-Cogito-7B    & 79.6 & 68.1 & 40.0 & 73.0 & 73.2 & 3.95 & 3.63 & 85.0 & 85.0 & 84.1 \\
        \rowcolor{gray!20}
        \textbf{+ LEAD (Ours)} & 81.7~\color{blue!60}{(+2.1)} & 69.2~\color{blue!60}{(+1.1)} & 42.0~\color{blue!60}{(+2.0)} & 75.6~\color{blue!60}{(+2.6)} & 75.6~\color{blue!60}{(+2.4)} & 4.13~\color{blue!60}{(+3.0)} & 3.80~\color{blue!60}{(+2.8)} & 86.3~\color{blue!60}{(+1.3)} & 86.6~\color{blue!60}{(+1.6)} & 86.1~\color{blue!60}{(+2.0)} \\

        \cdashline{1-11}
        OpenVLThinker-7B & 68.1 & 62.3 & 46.5 & 71.5 & 80.3 & 3.59 & 3.50 & 82.4 & 82.5 & 79.1 \\
        \rowcolor{gray!20}
        \textbf{+ LEAD (Ours)} &
        70.2~\color{blue!60}{(+2.1)} &
        65.3~\color{blue!60}{(+3.0)} &
        47.2~\color{blue!60}{(+0.7)} &
        73.5~\color{blue!60}{(+2.0)} &
        81.3~\color{blue!60}{(+1.0)} &
        3.76~\color{blue!60}{(+2.8)} &
        3.71~\color{blue!60}{(+4.2)} &
        84.1~\color{blue!60}{(+1.7)} &
        83.5~\color{blue!60}{(+1.0)} &
        80.2~\color{blue!60}{(+1.1)} \\
        \hline
    \end{tabular}}
\end{table*}

\begin{table*}[h]
    \centering
    \footnotesize
    \caption{Comparisons of different MLRMs with LEAD across mathematical and scientific visual reasoning benchmarks.}
    \vspace{-0.2cm}
    \label{tab:specific_results}
    \resizebox{\textwidth}{!}{
    \setlength{\tabcolsep}{4pt}
    \renewcommand{\arraystretch}{1.3}
    \begin{tabular}{l|lllllll|lll}
        \hline
        \rowcolor{mygray}
        \multicolumn{1}{l|}{\cellcolor{mygray}} &
        \multicolumn{7}{c|}{\cellcolor{mygray}\textbf{Mathematical Reasoning}} &
        \multicolumn{3}{c}{\cellcolor{mygray}\textbf{Scientific Reasoning}} \\
        \cline{2-11}
        \rowcolor{mygray}
        \multicolumn{1}{l|}{\multirow{-2}{*}{\cellcolor{mygray}\textbf{Method}}} &
        \textbf{MathVision}$\uparrow$ & \textbf{MathVista}$\uparrow$ & \textbf{MathVerse}$\uparrow$ &
        \textbf{VisuLogic}$\uparrow$ & \textbf{Geometry3K}$\uparrow$ & \textbf{MMK12-Math}$\uparrow$ &
        & \textbf{MMK12-Phys}$\uparrow$ & \textbf{MMK12-Chem}$\uparrow$ & \textbf{MMK12-Bio}$\uparrow$ \\
        \hline

        R1-Onevision-7B & 29.9 & 64.1 & 46.4 & 24.9 & 57.9 & 44.8 & & 33.8 & 39.8 & 40.8 \\
        \rowcolor{gray!20}
        \textbf{+ LEAD (Ours)} & 
        32.4~\color{blue!60}{(+2.5)} & 66.4~\color{blue!60}{(+2.3)} & 47.3~\color{blue!60}{(+0.9)} &
        26.1~\color{blue!60}{(+1.2)} & 61.2~\color{blue!60}{(+3.3)} & 46.7~\color{blue!60}{(+1.9)} & &
        36.1~\color{blue!60}{(+2.3)} & 43.2~\color{blue!60}{(+3.4)} & 44.8~\color{blue!60}{(+4.0)} \\
        \cdashline{1-11}
        
        Vision-R1-7B & 27.2 & 73.5 & 52.4 & 26.4 & 67.0 & 52.1 & & 47.3 & 55.4 & 57.9 \\
        \rowcolor{gray!20}
        \textbf{+ LEAD (Ours)} &
        29.7~\color{blue!60}{(+2.5)} & 74.9~\color{blue!60}{(+1.4)} & 54.5~\color{blue!60}{(+2.1)} &
        27.9~\color{blue!60}{(+1.5)} & 68.3~\color{blue!60}{(+1.3)} & 53.9~\color{blue!60}{(+1.8)} & &
        49.2~\color{blue!60}{(+1.9)} & 56.6~\color{blue!60}{(+1.2)} & 58.6~\color{blue!60}{(+0.7)} \\
        \cdashline{1-11}
        
        VL-Rethinker-7B & 32.3 & 74.9 & 54.2 & 27.3 & 67.7 & 51.3 & & 47.2 & 57.4 & 64.8 \\
        \rowcolor{gray!20}
        \textbf{+ LEAD (Ours)} &
        33.1~\color{blue!60}{(+0.8)} & 75.6~\color{blue!60}{(+0.7)} & 54.9~\color{blue!60}{(+0.7)} &
        28.5~\color{blue!60}{(+1.2)} & 68.9~\color{blue!60}{(+1.2)} & 52.4~\color{blue!60}{(+1.1)} & &
        49.1~\color{blue!60}{(+1.9)} & 60.6~\color{blue!60}{(+3.2)} & 65.6~\color{blue!60}{(+0.8)} \\
        \cdashline{1-11}
        
        VL-Cogito-7B & 30.7 & 74.8 & 53.3 & 28.2 & 68.7 & 63.7 & & 43.2 & 57.5 & 61.3 \\
        \rowcolor{gray!20}
        \textbf{+ LEAD (Ours)} &
        32.4~\color{blue!60}{(+1.7)} & 76.3~\color{blue!60}{(+1.5)} & 55.1~\color{blue!60}{(+1.8)} &
        28.9~\color{blue!60}{(+0.7)} & 69.1~\color{blue!60}{(+0.4)} & 65.1~\color{blue!60}{(+1.4)} & &
        44.6~\color{blue!60}{(+1.4)} & 58.4~\color{blue!60}{(+0.9)} & 64.6~\color{blue!60}{(+3.3)} \\
        \hline
    \end{tabular}}
    \vspace{-0.2cm}
\end{table*}

\vspace{-1em}

\begin{figure}
    \centering
    \includegraphics[width=\linewidth]{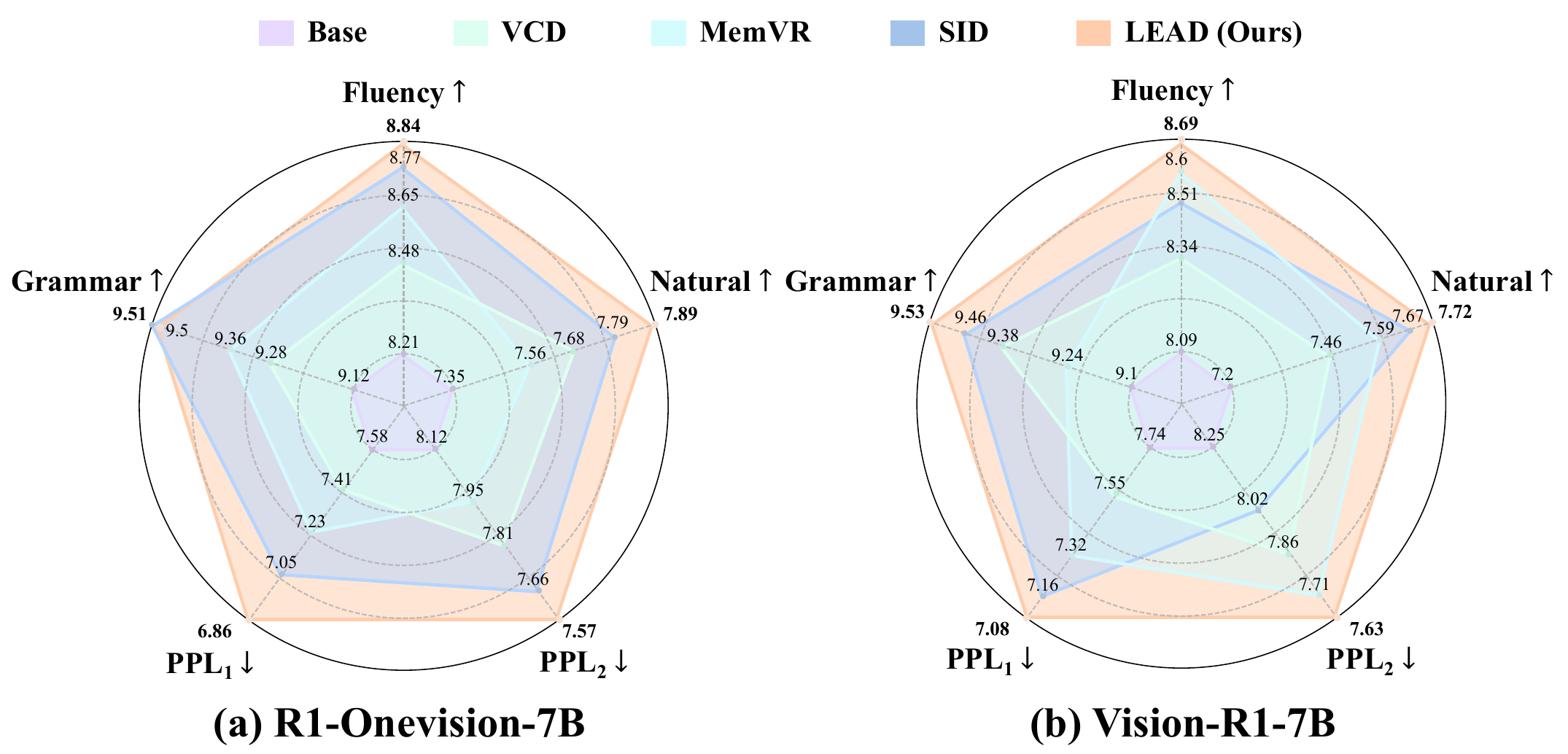}
    \vspace{-0.6cm}
    \caption{The average performance is evaluated on MMHalu using R1-Onevision-7B and Vision-R1-7B. PPL$_1$ and PPL$_2$ are calculated using gpt2, while the ratings for Grammar, Fluency and Naturalness are provided by GPT-5.}
    \label{fig:placeholder}
\end{figure}

\begin{figure}
    \centering
    \includegraphics[width=\linewidth]{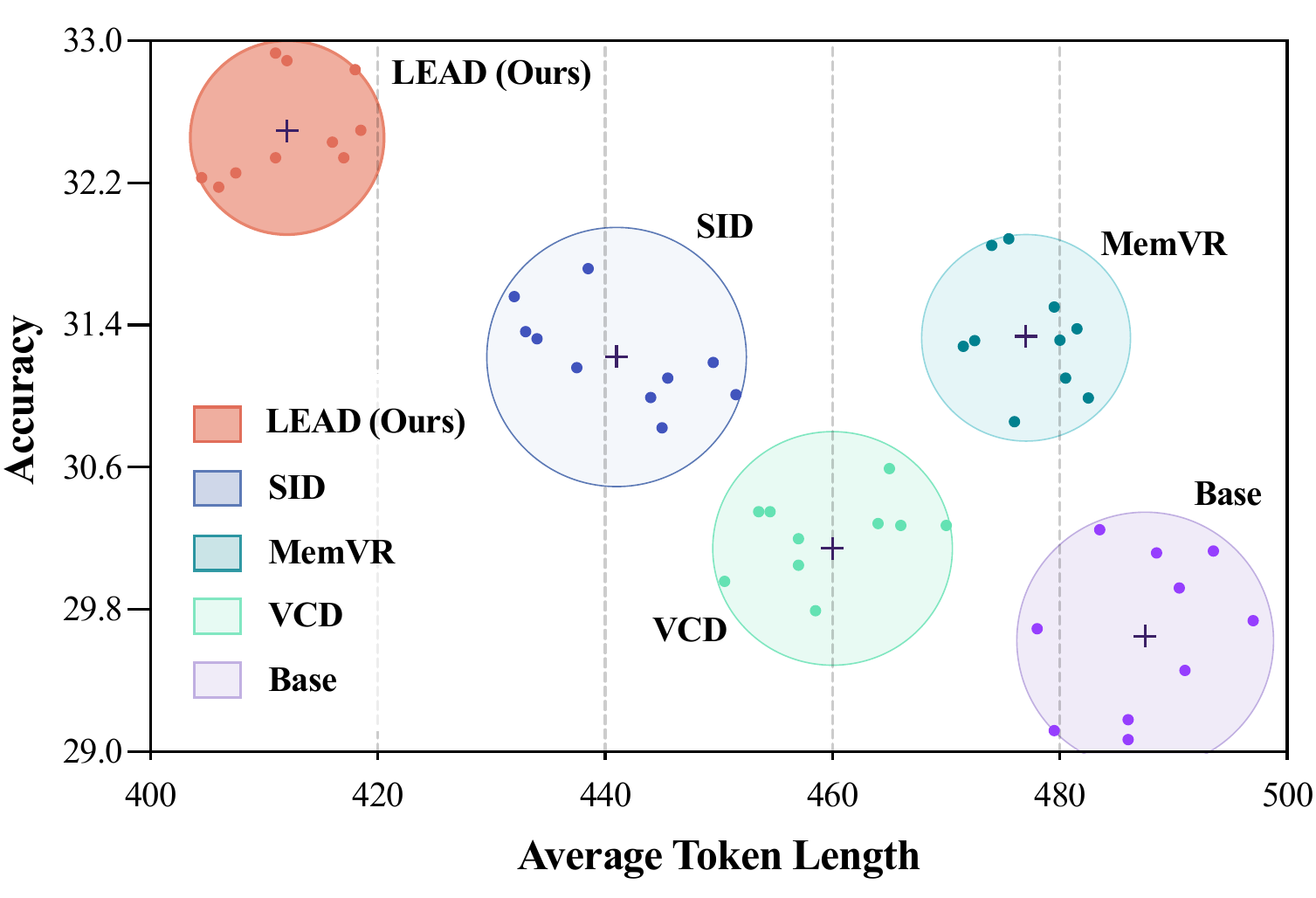}
    \vspace{-0.6cm}
    \caption{Comparisons of accuracy and reasoning length across multiple hallucination mitigation methods. The x-axis represents the average reasoning length computed on the MathVision dataset with R1-Onevision-7B.}
    \label{fig:thinking_efficiency}
    \vspace{-0.2cm}
\end{figure}

\begin{figure}
    \centering
    \includegraphics[width=\linewidth]{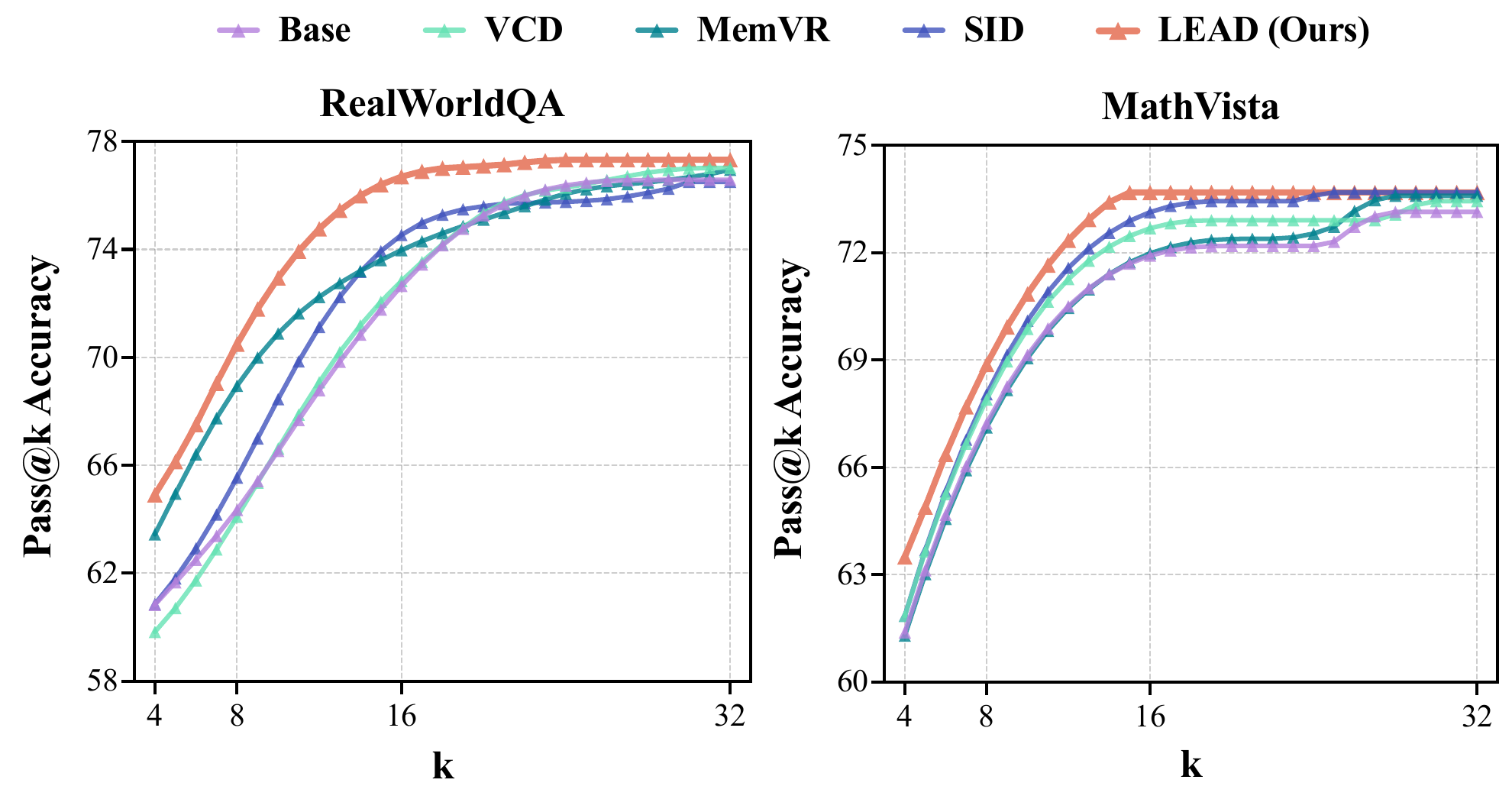}
    \vspace{-0.6cm}
    \caption{Pass@k accuracy evaluation of R1-Onevision-7B on sampled data of RealworldQA and MathVista, illustrating results for $k\in[4,32]$.}
    \label{fig:passk}
    \vspace{-0.2cm}
\end{figure}

\paragraph{GPT-5 Assisted Evaluation.}

To comprehensively assess the quality of the generated text, we employ the Perplexity (PPL) metric and utilize GPT-5 to evaluate grammar, fluency, and naturalness  of the text. We conduct evaluations on the MMHalu dataset using R1-OneVision and Vision-R1. As demonstrated in Fig.~\ref{fig:placeholder}, LEAD consistently preserves the quality of the generated text across multiple dimensions.

\vspace{-1em}

\paragraph{Reasoning Efficiency.}

We evaluate reasoning efficiency on MathVision using R1-Onevision, as shown in Fig.\ref{fig:thinking_efficiency}. LEAD generates shorter reasoning length than the baselines while maintaining the highest accuracy. This efficiency gain is attributed to the latent reasoning phase, which allows the model to retain multiple reasoning hypotheses at each step and reach the solutions with fewer generated tokens. 

\vspace{-1em}

\paragraph{Pass@k Performance.}

In addition to Pass@1, we evaluate Pass@k performance for $k\in[1,64]$ on R1-Onevision and compare it with other methods. We show results for $k\in[4,32]$ for better illustration (See Appendix C for the full results). As shown in Figure \ref{fig:passk}, LEAD reaches its peak accuracy at smaller $k$ values than the baselines, indicating higher sample efficiency. In addition to requiring fewer samples to reach peak accuracy, LEAD also shows a steeper increase in Pass@k at small $k$ and attains a higher final accuracy than VCD and MemVR. This indicates greater diversity of LEAD in reasoning and greater correctness. 

\section{Conclusion}
In this work, we examine token-level uncertainty and reveal that transition words frequently coincide with high-entropy reasoning states, which exhibit a strong association with hallucination-prone behaviors. Additionally, we find that high-entropy tokens linked to hallucinations tend to receive markedly lower visual attention, indicating that the model tends to overlook visual information under uncertainty. Motivated by these observations, we present LEAD, a lightweight and plug-and-play decoding framework that adaptively alternates between discrete and latent semantic representations, while incorporating visual guidance during high-uncertainty phases to enhance reasoning stability. Extensive evaluations on both general-purpose and scientific benchmarks demonstrate that LEAD consistently strengthens reasoning reliability and significantly reduces multimodal hallucinations.
\clearpage

\section*{Acknowledgments}
This research was supported by the Australian Government Research Training Program (RTP) Scholarship.

{
    \small
    \bibliographystyle{ieeenat_fullname}
    \bibliography{main}

@String(ECCV= {Eur. Conf. Comput. Vis.})

@String(ICLR = {Int. Conf. Learn. Represent.})

@String(AAAI = {AAAI})

@String(ECCV  = {ECCV})

@String(ICLR  = {ICLR})

@article{wang2024mathvision,
  title={Measuring multimodal mathematical reasoning with math-vision dataset},
  author={Wang, Ke and Pan, Junting and Shi, Weikang and Lu, Zimu and Ren, Houxing and Zhou, Aojun and Zhan, Mingjie and Li, Hongsheng},
  journal={Advances in Neural Information Processing Systems},
  volume={37},
  pages={95095--95169},
  year={2024}
}

@inproceedings{leng2024vcd,
  title={Mitigating object hallucinations in large vision-language models through visual contrastive decoding},
  author={Leng, Sicong and Zhang, Hang and Chen, Guanzheng and Li, Xin and Lu, Shijian and Miao, Chunyan and Bing, Lidong},
  booktitle={Proceedings of the IEEE/CVF Conference on Computer Vision and Pattern Recognition},
  pages={13872--13882},
  year={2024}
}

@article{zou2024look,
  title={Look twice before you answer: Memory-space visual retracing for hallucination mitigation in multimodal large language models},
  author={Zou, Xin and Wang, Yizhou and Yan, Yibo and Lyu, Yuanhuiyi and Zheng, Kening and Huang, Sirui and Chen, Junkai and Jiang, Peijie and Liu, Jia and Tang, Chang and others},
  journal={arXiv preprint arXiv:2410.03577},
  year={2024}
}

@article{guo2025deepseek,
  title={Deepseek-r1: Incentivizing reasoning capability in llms via reinforcement learning},
  author={Guo, Daya and Yang, Dejian and Zhang, Haowei and Song, Junxiao and Zhang, Ruoyu and Xu, Runxin and Zhu, Qihao and Ma, Shirong and Wang, Peiyi and Bi, Xiao and others},
  journal={arXiv preprint arXiv:2501.12948},
  year={2025}
}

@article{yang2025r1onevision,
  title={R1-onevision: Advancing generalized multimodal reasoning through cross-modal formalization},
  author={Yang, Yi and He, Xiaoxuan and Pan, Hongkun and Jiang, Xiyan and Deng, Yan and Yang, Xingtao and Lu, Haoyu and Yin, Dacheng and Rao, Fengyun and Zhu, Minfeng and others},
  journal={arXiv preprint arXiv:2503.10615},
  year={2025}
}

@inproceedings{li2023pope,
  title={Evaluating Object Hallucination in Large Vision-Language Models},
  author={Li, Yifan and Du, Yifan and Zhou, Kun and Wang, Jinpeng and Zhao, Wayne Xin and Wen, Ji-Rong},
  booktitle={Proceedings of the 2023 Conference on Empirical Methods in Natural Language Processing},
  pages={292--305},
  year={2023}
}

@inproceedings{sun2023aligning,
  title={Aligning Large Multimodal Models with Factually Augmented RLHF},
  author={Sun, Zhiqing and Shen, Sheng and Cao, Shengcao and Liu, Haotian and Li, Chunyuan and Shen, Yikang and Gan, Chuang and Gui, Liang-Yan and Wang, Yu-Xiong and Yang, Yiming and others},
  booktitle={Annual Meeting of the Association for Computational Linguistics},
  year={2024}
}

@article{dong2024insight,
  title={Insight-V: Exploring Long-Chain Visual Reasoning with Multimodal Large Language Models},
  author={Dong, Yuhao and Liu, Zuyan and Sun, Hai-Long and Yang, Jingkang and Hu, Winston and Rao, Yongming and Liu, Ziwei},
  journal={arXiv preprint arXiv:2411.14432},
  year={2024}
}

@article{lu2021inter,
  title={Inter-GPS: Interpretable geometry problem solving with formal language and symbolic reasoning},
  author={Lu, Pan and Gong, Ran and Jiang, Shibiao and Qiu, Liang and Huang, Siyuan and Liang, Xiaodan and Zhu, Song-Chun},
  journal={arXiv preprint arXiv:2105.04165},
  year={2021}
}

@inproceedings{zhang2024mathverse,
  title={Mathverse: Does your multi-modal llm truly see the diagrams in visual math problems?},
  author={Zhang, Renrui and Jiang, Dongzhi and Zhang, Yichi and Lin, Haokun and Guo, Ziyu and Qiu, Pengshuo and Zhou, Aojun and Lu, Pan and Chang, Kai-Wei and Qiao, Yu and others},
  booktitle={European Conference on Computer Vision},
  year={2024},
  organization={Springer}
}

@article{visionr1,
  title={Vision-r1: Incentivizing reasoning capability in multimodal large language models},
  author={Huang, Wenxuan and Jia, Bohan and Zhai, Zijie and Cao, Shaosheng and Ye, Zheyu and Zhao, Fei and Hu, Yao and Lin, Shaohui},
  journal={arXiv preprint arXiv:2503.06749},
  year={2025}
}

@article{lrm-openaio1,
  author  = {OpenAI},
  title   = {Learning to Reason with {LLMs}},
  year    = {2024},
  url     = {https://openai.com/index/learning-to-reason-with-llms/}
}

@article{lrm_demystifying,
  author       = {Edward Y. Chang and
                  Yuxuan Tong and
                  Morry Niu and
                  Graham Neubig and
                  Xiang Yue},
  title        = {Demystifying Long Chain-of-Thought Reasoning in LLMs},
  journal      = {CoRR},
  volume       = {abs/2502.03373},
  year         = {2025}
}

@article{meng2025mm,
  title={MM-Eureka: Exploring Visual Aha Moment with Rule-based Large-scale Reinforcement Learning},
  author={Meng, Fanqing and Du, Lingxiao and Liu, Zongkai and Zhou, Zhixiang and Lu, Quanfeng and Fu, Daocheng and Shi, Botian and Wang, Wenhai and He, Junjun and Zhang, Kaipeng and others},
  journal={arXiv preprint arXiv:2503.07365},
  year={2025}
}

@inproceedings{geometry3k,
    title = "{I}nter-{GPS}: Interpretable Geometry Problem Solving with Formal Language and Symbolic Reasoning",
    author = "Lu, Pan  and
      Gong, Ran  and
      Jiang, Shibiao  and
      Qiu, Liang  and
      Huang, Siyuan  and
      Liang, Xiaodan  and
      Zhu, Song-Chun",
    editor = "Zong, Chengqing  and
      Xia, Fei  and
      Li, Wenjie  and
      Navigli, Roberto",
    booktitle = "Proceedings of the 59th Annual Meeting of the Association for Computational Linguistics and the 11th International Joint Conference on Natural Language Processing (Volume 1: Long Papers)",
    month = aug,
    year = "2021",
    address = "Online",
    publisher = "Association for Computational Linguistics",
    url = "https://aclanthology.org/2021.acl-long.528/",
    doi = "10.18653/v1/2021.acl-long.528",
    pages = "6774--6786"
}

@inproceedings{
      mathvision,
      title={Measuring Multimodal Mathematical Reasoning with MATH-Vision Dataset},
      author={Ke Wang and Junting Pan and Weikang Shi and Zimu Lu and Houxing Ren and Aojun Zhou and Mingjie Zhan and Hongsheng Li},
      booktitle={The Thirty-eight Conference on Neural Information Processing Systems Datasets and Benchmarks Track},
      year={2024},
      url={https://openreview.net/forum?id=QWTCcxMpPA}
}

@InProceedings{mathverse,
author="Zhang, Renrui
and Jiang, Dongzhi
and Zhang, Yichi
and Lin, Haokun
and Guo, Ziyu
and Qiu, Pengshuo
and Zhou, Aojun
and Lu, Pan
and Chang, Kai-Wei
and Qiao, Yu
and Gao, Peng
and Li, Hongsheng",
editor="Leonardis, Ale{\v{s}}
and Ricci, Elisa
and Roth, Stefan
and Russakovsky, Olga
and Sattler, Torsten
and Varol, G{\"u}l",
title="MATHVERSE: Does Your Multi-modal LLM Truly See the Diagrams in Visual Math Problems?",
booktitle="Computer Vision -- ECCV 2024",
year="2025",
publisher="Springer Nature Switzerland",
address="Cham",
pages="169--186",
isbn="978-3-031-73242-3"
}

@inproceedings{mathvista,
  author    = {Lu, Pan and Bansal, Hritik and Xia, Tony and Liu, Jiacheng and Li, Chunyuan and Hajishirzi, Hannaneh and Cheng, Hao and Chang, Kai-Wei and Galley, Michel and Gao, Jianfeng},
  title     = {MathVista: Evaluating Mathematical Reasoning of Foundation Models in Visual Contexts},
  booktitle={International Conference on Learning Representations (ICLR)},
  year      = {2024}
}

@article{xu2025visulogic,
  title={Visulogic: A benchmark for evaluating visual reasoning in multi-modal large language models},
  author={Xu, Weiye and Wang, Jiahao and Wang, Weiyun and Chen, Zhe and Zhou, Wengang and Yang, Aijun and Lu, Lewei and Li, Houqiang and Wang, Xiaohua and Zhu, Xizhou and others},
  journal={arXiv preprint arXiv:2504.15279},
  year={2025}
}

@misc{ming2025oceanr1,
  author       = {Lingfeng, Ming and Yadong, Li and Song, Chen and Jianhua, Xu and Zenan, Zhou and Weipeng, Chen},
  title        = {Ocean-R1: An Open and Generalizable Large Vision-Language Model enhanced by Reinforcement Learning},
  note         = {Accessed: 2025-04-03},
  year         = {2025}
}

@inproceedings{tong2024eyes,
  title={Eyes wide shut? exploring the visual shortcomings of multimodal llms},
  author={Tong, Shengbang and Liu, Zhuang and Zhai, Yuexiang and Ma, Yi and LeCun, Yann and Xie, Saining},
  booktitle={Proceedings of the IEEE/CVF Conference on Computer Vision and Pattern Recognition},
  pages={9568--9578},
  year={2024}
}

@article{huang2024mmevalpro,
  title={Mmevalpro: Calibrating multimodal benchmarks towards trustworthy and efficient evaluation},
  author={Huang, Jinsheng and Chen, Liang and Guo, Taian and Zeng, Fu and Zhao, Yusheng and Wu, Bohan and Yuan, Ye and Zhao, Haozhe and Guo, Zhihui and Zhang, Yichi and others},
  journal={arXiv preprint arXiv:2407.00468},
  year={2024}
}

@article{zhang2025automated,
  title={Automated Generation of Challenging Multiple-Choice Questions for Vision Language Model Evaluation},
  author={Zhang, Yuhui and Su, Yuchang and Liu, Yiming and Wang, Xiaohan and Burgess, James and Sui, Elaine and Wang, Chenyu and Aklilu, Josiah and Lozano, Alejandro and Wei, Anjiang and others},
  journal={arXiv preprint arXiv:2501.03225},
  year={2025}
}

@article{cui2023holistic,
  title={Holistic analysis of hallucination in gpt-4v (ision): Bias and interference challenges},
  author={Cui, Chenhang and Zhou, Yiyang and Yang, Xinyu and Wu, Shirley and Zhang, Linjun and Zou, James and Yao, Huaxiu},
  journal={arXiv preprint arXiv:2311.03287},
  year={2023}
}

@article{wang2025sota,
  title={SoTA with Less: MCTS-Guided Sample Selection for Data-Efficient Visual Reasoning Self-Improvement},
  author={Wang, Xiyao and Yang, Zhengyuan and Feng, Chao and Lu, Hongjin and Li, Linjie and Lin, Chung-Ching and Lin, Kevin and Huang, Furong and Wang, Lijuan},
  journal={arXiv preprint arXiv:2504.07934},
  year={2025}
}

@inproceedings{wu2024v,
  title={V?: Guided visual search as a core mechanism in multimodal llms},
  author={Wu, Penghao and Xie, Saining},
  booktitle={Proceedings of the IEEE/CVF Conference on Computer Vision and Pattern Recognition},
  pages={13084--13094},
  year={2024}
}

@article{liu2025visual,
  author = {Liu, Ziyu and Sun, Zeyi and Zang, Yuhang and Dong, Xiaoyi and Cao, Yuhang and Duan, Haodong and Lin, Dahua and Wang, Jiaqi},
  journal = {arXiv preprint arXiv:2503.01785},
  title = {Visual-RFT: Visual Reinforcement Fine-Tuning},
  year = {2025}
}

@misc{liu2025segzero,
  archiveprefix = {arXiv},
  author = {Yuqi, Liu and Bohao, Peng and Zhisheng, Zhong and Zihao, Yue and Fanbin, Lu and Bei, Yu and Jiaya, Jia},
  eprint = {2503.06520},
  primaryclass = {cs.CV},
  title = {Seg-Zero: Reasoning-Chain Guided Segmentation via Cognitive Reinforcement},
  url = {https://arxiv.org/abs/2503.06520},
  year = {2025}
}

@article{xiao2025fast,
  title={Fast-Slow Thinking for Large Vision-Language Model Reasoning},
  author={Xiao, Wenyi and Gan, Leilei and Dai, Weilong and He, Wanggui and Huang, Ziwei and Li, Haoyuan and Shu, Fangxun and Yu, Zhelun and Zhang, Peng and Jiang, Hao and others},
  journal={arXiv preprint arXiv:2504.18458},
  year={2025}
}

@article{wang2025vl,
  title={VL-Rethinker: Incentivizing Self-Reflection of Vision-Language Models with Reinforcement Learning},
  author={Wang, Haozhe and Qu, Chao and Huang, Zuming and Chu, Wei and Lin, Fangzhen and Chen, Wenhu},
  journal={arXiv preprint arXiv:2504.08837},
  year={2025}
}

@article{tan2025reason,
  title={Reason-rft: Reinforcement fine-tuning for visual reasoning},
  author={Tan, Huajie and Ji, Yuheng and Hao, Xiaoshuai and Lin, Minglan and Wang, Pengwei and Wang, Zhongyuan and Zhang, Shanghang},
  journal={arXiv preprint arXiv:2503.20752},
  year={2025}
}

@article{wang2025visualprm,
  title={Visualprm: An effective process reward model for multimodal reasoning},
  author={Wang, Weiyun and Gao, Zhangwei and Chen, Lianjie and Chen, Zhe and Zhu, Jinguo and Zhao, Xiangyu and Liu, Yangzhou and Cao, Yue and Ye, Shenglong and Zhu, Xizhou and others},
  journal={arXiv preprint arXiv:2503.10291},
  year={2025}
}

@article{zhang2025r1,
  title={R1-vl: Learning to reason with multimodal large language models via step-wise group relative policy optimization},
  author={Zhang, Jingyi and Huang, Jiaxing and Yao, Huanjin and Liu, Shunyu and Zhang, Xikun and Lu, Shijian and Tao, Dacheng},
  journal={arXiv preprint arXiv:2503.12937},
  year={2025}
}

@inproceedings{huang2024opera,
  title={Opera: Alleviating hallucination in multi-modal large language models via over-trust penalty and retrospection-allocation},
  author={Huang, Qidong and Dong, Xiaoyi and Zhang, Pan and Wang, Bin and He, Conghui and Wang, Jiaqi and Lin, Dahua and Zhang, Weiming and Yu, Nenghai},
  booktitle={Proceedings of the IEEE/CVF Conference on Computer Vision and Pattern Recognition},
  pages={13418--13427},
  year={2024}
}

@article{liuyue_GuardReasoner-VL,
  title={GuardReasoner-VL: Safeguarding VLMs via Reinforced Reasoning},
  author={Liu, Yue and Zhai, Shengfang and Du, Mingzhe and Chen, Yulin and Cao, Tri and Gao, Hongcheng and Wang, Cheng and Li, Xinfeng and Wang, Kun and Fang, Junfeng and Zhang, Jiaheng and Hooi, Bryan},
  journal={arXiv preprint arXiv:2505.11049},
  year={2025}
}

@article{dong2025mirage,
  title={MIRAGE: Assessing Hallucination in Multimodal Reasoning Chains of MLLM},
  author={Dong, Bowen and Ni, Minheng and Huang, Zitong and Yang, Guanglei and Zuo, Wangmeng and Zhang, Lei},
  journal={arXiv preprint arXiv:2505.24238},
  year={2025}
}

@article{tian2025more,
  title={More Thought, Less Accuracy? On the Dual Nature of Reasoning in Vision-Language Models},
  author={Tian, Xinyu and Zou, Shu and Yang, Zhaoyuan and He, Mengqi and Waschkowski, Fabian and Wesemann, Lukas and Tu, Peter and Zhang, Jing},
  journal={arXiv preprint arXiv:2509.25848},
  year={2025}
}

@article{song2025hallucination,
  title={The hallucination tax of reinforcement finetuning},
  author={Song, Linxin and Shi, Taiwei and Zhao, Jieyu},
  journal={arXiv preprint arXiv:2505.13988},
  year={2025}
}

@article{liu2025more,
  title={More Thinking, Less Seeing? Assessing Amplified Hallucination in Multimodal Reasoning Models},
  author={Liu, Chengzhi and Xu, Zhongxing and Wei, Qingyue and Wu, Juncheng and Zou, James and Wang, Xin Eric and Zhou, Yuyin and Liu, Sheng},
  journal={arXiv preprint arXiv:2505.21523},
  year={2025}
}

@article{cheng2025chain,
  title={Chain-of-Thought Prompting Obscures Hallucination Cues in Large Language Models: An Empirical Evaluation},
  author={Cheng, Jiahao and Su, Tiancheng and Yuan, Jia and He, Guoxiu and Liu, Jiawei and Tao, Xinqi and Xie, Jingwen and Li, Huaxia},
  journal={arXiv preprint arXiv:2506.17088},
  year={2025}
}

@article{lu2025auditing,
  title={Auditing Meta-Cognitive Hallucinations in Reasoning Large Language Models},
  author={Lu, Haolang and Liu, Yilian and Xu, Jingxin and Nan, Guoshun and Yu, Yuanlong and Chen, Zhican and Wang, Kun},
  journal={arXiv preprint arXiv:2505.13143},
  year={2025}
}

@article{li2025hallucination,
  title={The Hallucination Dilemma: Factuality-Aware Reinforcement Learning for Large Reasoning Models},
  author={Li, Junyi and Ng, Hwee Tou},
  journal={arXiv preprint arXiv:2505.24630},
  year={2025}
}

@article{sun2025detection,
  title={Detection and Mitigation of Hallucination in Large Reasoning Models: A Mechanistic Perspective},
  author={Sun, Zhongxiang and Wang, Qipeng and Wang, Haoyu and Zhang, Xiao and Xu, Jun},
  journal={arXiv preprint arXiv:2505.12886},
  year={2025}
}

@article{yao2025reasoning,
  title={Are Reasoning Models More Prone to Hallucination?},
  author={Yao, Zijun and Liu, Yantao and Chen, Yanxu and Chen, Jianhui and Fang, Junfeng and Hou, Lei and Li, Juanzi and Chua, Tat-Seng},
  journal={arXiv preprint arXiv:2505.23646},
  year={2025}
}

@article{lu2025mitigating,
  title={Mitigating Hallucination in Multimodal Reasoning via Functional Attention Control},
  author={Lu, Haolang and Chu, Bolun and Fu, WeiYe and Nan, Guoshun and Liu, Junning and Pan, Minghui and Li, Qiankun and Yu, Yi and Wang, Hua and Wang, Kun},
  journal={arXiv preprint arXiv:2510.10285},
  year={2025}
}

@article{xiao2025advancing,
  title={Advancing Multimodal Reasoning Capabilities of Multimodal Large Language Models via Visual Perception Reward},
  author={Xiao, Tong and Xu, Xin and Huang, Zhenya and Gao, Hongyu and Liu, Quan and Liu, Qi and Chen, Enhong},
  journal={arXiv preprint arXiv:2506.07218},
  year={2025}
}

@article{yu2025perception,
  title={Perception-r1: Pioneering perception policy with reinforcement learning},
  author={Yu, En and Lin, Kangheng and Zhao, Liang and Yin, Jisheng and Wei, Yana and Peng, Yuang and Wei, Haoran and Sun, Jianjian and Han, Chunrui and Ge, Zheng and others},
  journal={arXiv preprint arXiv:2504.07954},
  year={2025}
}

@article{ding2025vtperception,
  title={VTPerception-R1: Enhancing Multimodal Reasoning via Explicit Visual and Textual Perceptual Grounding},
  author={Ding, Yizhuo and Chen, Mingkang and Feng, Zhibang and Xiao, Tong and Qu, Wanying and Shao, Wenqi and Fu, Yanwei},
  journal={arXiv preprint arXiv:2509.24776},
  year={2025}
}

@article{wang2025perception,
  title={Perception-aware policy optimization for multimodal reasoning},
  author={Wang, Zhenhailong and Guo, Xuehang and Stoica, Sofia and Xu, Haiyang and Wang, Hongru and Ha, Hyeonjeong and Chen, Xiusi and Chen, Yangyi and Yan, Ming and Huang, Fei and others},
  journal={arXiv preprint arXiv:2507.06448},
  year={2025}
}

@article{wei2025open,
  title={Open vision reasoner: Transferring linguistic cognitive behavior for visual reasoning},
  author={Wei, Yana and Zhao, Liang and Sun, Jianjian and Lin, Kangheng and Yin, Jisheng and Hu, Jingcheng and Zhang, Yinmin and Yu, En and Lv, Haoran and Weng, Zejia and others},
  journal={arXiv preprint arXiv:2507.05255},
  year={2025}
}

@article{chung2025mllms,
  title={What MLLMs Learn about When they Learn about Multimodal Reasoning: Perception, Reasoning, or their Integration?},
  author={Chung, Jiwan and Joshi, Neel and Sharma, Pratyusha and Yu, Youngjae and Vineet, Vibhav},
  journal={arXiv preprint arXiv:2510.01719},
  year={2025}
}

@article{li2025mixture,
  title={Mixture-of-Visual-Thoughts: Exploring Context-Adaptive Reasoning Mode Selection for General Visual Reasoning},
  author={Li, Zejun and Zhao, Yingxiu and Zhang, Jiwen and Wang, Siyuan and Yao, Yang and Zhao, Runzhou and Song, Jun and Zheng, Bo and Wei, Zhongyu},
  journal={arXiv preprint arXiv:2509.22746},
  year={2025}
}

@article{leng2025mmr1,
  title={MMR1: Enhancing Multimodal Reasoning with Variance-Aware Sampling and Open Resources},
  author={Leng, Sicong and Wang, Jing and Li, Jiaxi and Zhang, Hao and Hu, Zhiqiang and Zhang, Boqiang and Jiang, Yuming and Zhang, Hang and Li, Xin and Bing, Lidong and others},
  journal={arXiv preprint arXiv:2509.21268},
  year={2025}
}

@article{wang2025internvl3,
  title={Internvl3. 5: Advancing open-source multimodal models in versatility, reasoning, and efficiency},
  author={Wang, Weiyun and Gao, Zhangwei and Gu, Lixin and Pu, Hengjun and Cui, Long and Wei, Xingguang and Liu, Zhaoyang and Jing, Linglin and Ye, Shenglong and Shao, Jie and others},
  journal={arXiv preprint arXiv:2508.18265},
  year={2025}
}

@article{zhang2025thyme,
  title={Thyme: Think beyond images},
  author={Zhang, Yi-Fan and Lu, Xingyu and Yin, Shukang and Fu, Chaoyou and Chen, Wei and Hu, Xiao and Wen, Bin and Jiang, Kaiyu and Liu, Changyi and Zhang, Tianke and others},
  journal={arXiv preprint arXiv:2508.11630},
  year={2025}
}

@article{liang2025mm,
  title={MM-R1: Unleashing the Power of Unified Multimodal Large Language Models for Personalized Image Generation},
  author={Liang, Qian and Wu, Yujia and Li, Kuncheng and Wei, Jiwei and He, Shiyuan and Guo, Jinyu and Xie, Ning},
  journal={arXiv preprint arXiv:2508.11433},
  year={2025}
}

@article{qiao2025we,
  title={We-math 2.0: A versatile mathbook system for incentivizing visual mathematical reasoning},
  author={Qiao, Runqi and Tan, Qiuna and Yang, Peiqing and Wang, Yanzi and Wang, Xiaowan and Wan, Enhui and Zhou, Sitong and Dong, Guanting and Zeng, Yuchen and Xu, Yida and others},
  journal={arXiv preprint arXiv:2508.10433},
  year={2025}
}

@article{xiao2025m2io,
  title={M2io-r1: An efficient rl-enhanced reasoning framework for multimodal retrieval augmented multimodal generation},
  author={Xiao, Zhiyou and Yu, Qinhan and Li, Binghui and Chen, Geng and Chen, Chong and Zhang, Wentao},
  journal={arXiv preprint arXiv:2508.06328},
  year={2025}
}

@article{chen2025sifthinker,
  title={Sifthinker: Spatially-aware image focus for visual reasoning},
  author={Chen, Zhangquan and Zhao, Ruihui and Luo, Chuwei and Sun, Mingze and Yu, Xinlei and Kang, Yangyang and Huang, Ruqi},
  journal={arXiv preprint arXiv:2508.06259},
  year={2025}
}

@article{zhu2025shuffle,
  title={Shuffle-r1: Efficient rl framework for multimodal large language models via data-centric dynamic shuffle},
  author={Zhu, Linghao and Guan, Yiran and Liang, Dingkang and Ju, Jianzhong and Luo, Zhenbo and Qin, Bin and Luan, Jian and Liu, Yuliang and Bai, Xiang},
  journal={arXiv preprint arXiv:2508.05612},
  year={2025}
}

@article{chen2025learning,
  title={Learning only with images: Visual reinforcement learning with reasoning, rendering, and visual feedback},
  author={Chen, Yang and Shen, Yufan and Huang, Wenxuan and Zhou, Sheng and Lin, Qunshu and Cai, Xinyu and Yu, Zhi and Bu, Jiajun and Shi, Botian and Qiao, Yu},
  journal={arXiv preprint arXiv:2507.20766},
  year={2025}
}

@article{ni2025point,
  title={Point-rft: Improving multimodal reasoning with visually grounded reinforcement finetuning},
  author={Ni, Minheng and Yang, Zhengyuan and Li, Linjie and Lin, Chung-Ching and Lin, Kevin and Zuo, Wangmeng and Wang, Lijuan},
  journal={arXiv preprint arXiv:2505.19702},
  year={2025}
}

@article{wang2025vrag,
  title={VRAG-RL: Empower Vision-Perception-Based RAG for Visually Rich Information Understanding via Iterative Reasoning with Reinforcement Learning},
  author={Wang, Qiuchen and Ding, Ruixue and Zeng, Yu and Chen, Zehui and Chen, Lin and Wang, Shihang and Xie, Pengjun and Huang, Fei and Zhao, Feng},
  journal={arXiv preprint arXiv:2505.22019},
  year={2025}
}

@article{wei2025advancing,
  title={Advancing Multimodal Reasoning via Reinforcement Learning with Cold Start},
  author={Wei, Lai and Li, Yuting and Zheng, Kaipeng and Wang, Chen and Wang, Yue and Kong, Linghe and Sun, Lichao and Huang, Weiran},
  journal={arXiv preprint arXiv:2505.22334},
  year={2025}
}

@inproceedings{weifirst,
  title={First SFT, Second RL, Third UPT: Continual Improving Multi-Modal LLM Reasoning via Unsupervised Post-Training},
  author={Wei, Lai and Li, Yuting and Wang, Chen and Wang, Yue and Kong, Linghe and Huang, Weiran and Sun, Lichao},
  booktitle={The Thirty-ninth Annual Conference on Neural Information Processing Systems}
}

@article{mao2025unirl,
  title={UniRL: Self-Improving Unified Multimodal Models via Supervised and Reinforcement Learning},
  author={Mao, Weijia and Yang, Zhenheng and Shou, Mike Zheng},
  journal={arXiv preprint arXiv:2505.23380},
  year={2025}
}

@article{liang2025modomodo,
  title={MoDoMoDo: Multi-Domain Data Mixtures for Multimodal LLM Reinforcement Learning},
  author={Liang, Yiqing and Qiu, Jielin and Ding, Wenhao and Liu, Zuxin and Tompkin, James and Xu, Mengdi and Xia, Mengzhou and Tu, Zhengzhong and Shi, Laixi and Zhu, Jiacheng},
  journal={arXiv preprint arXiv:2505.24871},
  year={2025}
}

@article{wan2025srpo,
  title={Srpo: Enhancing multimodal llm reasoning via reflection-aware reinforcement learning},
  author={Wan, Zhongwei and Dou, Zhihao and Liu, Che and Zhang, Yu and Cui, Dongfei and Zhao, Qinjian and Shen, Hui and Xiong, Jing and Xin, Yi and Jiang, Yifan and others},
  journal={arXiv preprint arXiv:2506.01713},
  year={2025}
}

@article{wu2025synthrl,
  title={SynthRL: Scaling Visual Reasoning with Verifiable Data Synthesis},
  author={Wu, Zijian and Ni, Jinjie and Liu, Xiangyan and Liu, Zichen and Yan, Hang and Shieh, Michael Qizhe},
  journal={arXiv preprint arXiv:2506.02096},
  year={2025}
}

@article{chen2025advancing,
  title={Advancing Multimodal Reasoning: From Optimized Cold Start to Staged Reinforcement Learning},
  author={Chen, Shuang and Guo, Yue and Su, Zhaochen and Li, Yafu and Wu, Yulun and Chen, Jiacheng and Chen, Jiayu and Wang, Weijie and Qu, Xiaoye and Cheng, Yu},
  journal={arXiv preprint arXiv:2506.04207},
  year={2025}
}

@article{wu2025reinforcing,
  title={Reinforcing spatial reasoning in vision-language models with interwoven thinking and visual drawing},
  author={Wu, Junfei and Guan, Jian and Feng, Kaituo and Liu, Qiang and Wu, Shu and Wang, Liang and Wu, Wei and Tan, Tieniu},
  journal={arXiv preprint arXiv:2506.09965},
  year={2025}
}

@article{wang2025vicrit,
  title={ViCrit: A Verifiable Reinforcement Learning Proxy Task for Visual Perception in VLMs},
  author={Wang, Xiyao and Yang, Zhengyuan and Feng, Chao and Liang, Yongyuan and Zhou, Yuhang and Liu, Xiaoyu and Zang, Ziyi and Li, Ming and Lin, Chung-Ching and Lin, Kevin and others},
  journal={arXiv preprint arXiv:2506.10128},
  year={2025}
}

@article{hong2025glm,
  title={Glm-4.1 v-thinking: Towards versatile multimodal reasoning with scalable reinforcement learning},
  author={Hong, Wenyi and Yu, Wenmeng and Gu, Xiaotao and Wang, Guo and Gan, Guobing and Tang, Haomiao and Cheng, Jiale and Qi, Ji and Ji, Junhui and Pan, Lihang and others},
  journal={arXiv e-prints},
  pages={arXiv--2507},
  year={2025}
}

@article{chen2025synergy,
  title={The Synergy Dilemma of Long-CoT SFT and RL: Investigating Post-Training Techniques for Reasoning VLMs},
  author={Chen, Jierun and Yu, Tiezheng and Bai, Haoli and Yao, Lewei and Wu, Jiannan and Li, Kaican and Mi, Fei and Tao, Chaofan and Zhu, Lei and Zhang, Manyi and others},
  journal={arXiv preprint arXiv:2507.07562},
  year={2025}
}

@article{ai2025m2,
  title={M2-reasoning: Empowering mllms with unified general and spatial reasoning},
  author={AI, Inclusion and Wang, Fudong and Liu, Jiajia and Chen, Jingdong and Zhou, Jun and Ji, Kaixiang and Ru, Lixiang and Guo, Qingpei and Zheng, Ruobing and Li, Tianqi and others},
  journal={arXiv preprint arXiv:2507.08306},
  year={2025}
}

@inproceedings{leng2024mitigating,
  title={Mitigating object hallucinations in large vision-language models through visual contrastive decoding},
  author={Leng, Sicong and Zhang, Hang and Chen, Guanzheng and Li, Xin and Lu, Shijian and Miao, Chunyan and Bing, Lidong},
  booktitle={Proceedings of the IEEE/CVF Conference on Computer Vision and Pattern Recognition},
  pages={13872--13882},
  year={2024}
}

@article{wang2024mitigating,
  title={Mitigating hallucinations in large vision-language models with instruction contrastive decoding},
  author={Wang, Xintong and Pan, Jingheng and Ding, Liang and Biemann, Chris},
  journal={arXiv preprint arXiv:2403.18715},
  year={2024}
}

@article{huo2024self,
  title={Self-Introspective Decoding: Alleviating Hallucinations for Large Vision-Language Models},
  author={Huo, Fushuo and Xu, Wenchao and Zhang, Zhong and Wang, Haozhao and Chen, Zhicheng and Zhao, Peilin},
  journal={arXiv preprint arXiv:2408.02032},
  year={2024}
}

@article{liu2024paying,
  title={Paying More Attention to Image: A Training-Free Method for Alleviating Hallucination in LVLMs},
  author={Liu, Shi and Zheng, Kecheng and Chen, Wei},
  journal={arXiv preprint arXiv:2407.21771},
  year={2024}
}

@article{xing2024mitigating,
  title={Mitigating Object Hallucination via Concentric Causal Attention},
  author={Xing, Yun and Li, Yiheng and Laptev, Ivan and Lu, Shijian},
  journal={arXiv preprint arXiv:2410.15926},
  year={2024}
}

@inproceedings{ma2024vista,
  title={VISTA-LLAMA: Reducing Hallucination in Video Language Models via Equal Distance to Visual Tokens},
  author={Ma, Fan and Jin, Xiaojie and Wang, Heng and Xian, Yuchen and Feng, Jiashi and Yang, Yi},
  booktitle={Proceedings of the IEEE/CVF Conference on Computer Vision and Pattern Recognition},
  pages={13151--13160},
  year={2024}
}

@article{zhang2025self,
  title={Self-correcting decoding with generative feedback for mitigating hallucinations in large vision-language models},
  author={Zhang, Ce and Wan, Zifu and Kan, Zhehan and Ma, Martin Q and Stepputtis, Simon and Ramanan, Deva and Salakhutdinov, Russ and Morency, Louis-Philippe and Sycara, Katia and Xie, Yaqi},
  journal={arXiv preprint arXiv:2502.06130},
  year={2025}
}

@inproceedings{yin2025clearsight,
  title={ClearSight: Visual Signal Enhancement for Object Hallucination Mitigation in Multimodal Large Language Models},
  author={Yin, Hao and Si, Guangzong and Wang, Zilei},
  booktitle={Proceedings of the Computer Vision and Pattern Recognition Conference},
  year={2025}
}

@article{yuan2025vl,
  title={Vl-cogito: Progressive curriculum reinforcement learning for advanced multimodal reasoning},
  author={Yuan, Ruifeng and Xiao, Chenghao and Leng, Sicong and Wang, Jianyu and Li, Long and Xu, Weiwen and Chan, Hou Pong and Zhao, Deli and Xu, Tingyang and Wei, Zhongyu and others},
  journal={arXiv preprint arXiv:2507.22607},
  year={2025}
}

@inproceedings{deng2025openvlthinker,
  title={Openvlthinker: Complex vision-language reasoning via iterative sft-rl cycles},
  author={Deng, Yihe and Bansal, Hritik and Yin, Fan and Peng, Nanyun and Wang, Wei and Chang, Kai-Wei},
  booktitle={The Thirty-ninth Annual Conference on Neural Information Processing Systems},
  year={2025}
}

@article{cheng2025reasoning,
  title={Reasoning with exploration: An entropy perspective},
  author={Cheng, Daixuan and Huang, Shaohan and Zhu, Xuekai and Dai, Bo and Zhao, Wayne Xin and Zhang, Zhenliang and Wei, Furu},
  journal={arXiv preprint arXiv:2506.14758},
  year={2025}
}

@article{wang2025beyond,
  title={Beyond the 80/20 rule: High-entropy minority tokens drive effective reinforcement learning for llm reasoning},
  author={Wang, Shenzhi and Yu, Le and Gao, Chang and Zheng, Chujie and Liu, Shixuan and Lu, Rui and Dang, Kai and Chen, Xionghui and Yang, Jianxin and Zhang, Zhenru and others},
  journal={arXiv preprint arXiv:2506.01939},
  year={2025}
}

@article{qian2025demystifying,
  title={Demystifying reasoning dynamics with mutual information: Thinking tokens are information peaks in llm reasoning},
  author={Qian, Chen and Liu, Dongrui and Wen, Haochen and Bai, Zhen and Liu, Yong and Shao, Jing},
  journal={arXiv preprint arXiv:2506.02867},
  year={2025}
}

@article{huang2025pear,
  title={PEAR: Phase Entropy Aware Reward for Efficient Reasoning},
  author={Huang, Chen and Lu, Wei and Zhang, Wenxuan},
  journal={arXiv preprint arXiv:2510.08026},
  year={2025}
}

@article{sun2025latent,
  title={Latent Chain-of-Thought for Visual Reasoning},
  author={Sun, Guohao and Hua, Hang and Wang, Jian and Luo, Jiebo and Dianat, Sohail and Rabbani, Majid and Rao, Raghuveer and Tao, Zhiqiang},
  journal={arXiv preprint arXiv:2510.23925},
  year={2025}
}

@article{zhang2025soft,
  title={Soft thinking: Unlocking the reasoning potential of llms in continuous concept space},
  author={Zhang, Zhen and He, Xuehai and Yan, Weixiang and Shen, Ao and Zhao, Chenyang and Wang, Shuohang and Shen, Yelong and Wang, Xin Eric},
  journal={arXiv preprint arXiv:2505.15778},
  year={2025}
}

@inproceedings{zhuangmixture,
  title={Mixture of Inputs: Text Generation Beyond Discrete Token Sampling},
  author={Zhuang, Yufan and Liu, Liyuan and Singh, Chandan and Shang, Jingbo and Gao, Jianfeng},
  booktitle={The Thirty-ninth Annual Conference on Neural Information Processing Systems}
}

@article{wu2025llms,
  title={Llms are single-threaded reasoners: Demystifying the working mechanism of soft thinking},
  author={Wu, Junhong and Lu, Jinliang and Ren, Zixuan and Hu, Gangqiang and Wu, Zhi and Dai, Dai and Wu, Hua},
  journal={arXiv preprint arXiv:2508.03440},
  year={2025}
}

@article{shi2025swireasoning,
  title={SwiReasoning: Switch-Thinking in Latent and Explicit for Pareto-Superior Reasoning LLMs},
  author={Shi, Dachuan and Asi, Abedelkadir and Li, Keying and Yuan, Xiangchi and Pan, Leyan and Lee, Wenke and Xiao, Wen},
  journal={arXiv preprint arXiv:2510.05069},
  year={2025}
}

@article{hao2024training,
  title={Training large language models to reason in a continuous latent space},
  author={Hao, Shibo and Sukhbaatar, Sainbayar and Su, DiJia and Li, Xian and Hu, Zhiting and Weston, Jason and Tian, Yuandong},
  journal={arXiv preprint arXiv:2412.06769},
  year={2024}
}

@article{deng2025latent,
  title={Latent Reasoning in LLMs as a Vocabulary-Space Superposition},
  author={Deng, Jingcheng and Pang, Liang and Wei, Zihao and Xu, Shichen and Duan, Zenghao and Xu, Kun and Song, Yang and Shen, Huawei and Cheng, Xueqi},
  journal={arXiv preprint arXiv:2510.15522},
  year={2025}
}

@misc{grok2beta2024,
  title        = {Grok-2 Beta Release},
  author       = {{X.AI}},
  year         = {2024},
  note         = {Accessed: 2024}
}

@inproceedings{xu2025toward,
  title={Toward modality gap: Vision prototype learning for weakly-supervised semantic segmentation with clip},
  author={Xu, Zhongxing and Tang, Feilong and Chen, Zhe and Su, Yingxue and Zhao, Zhiyi and Zhang, Ge and Su, Jionglong and Ge, Zongyuan},
  booktitle={Proceedings of the AAAI Conference on Artificial Intelligence},
  volume={39},
  number={9},
  pages={9023--9031},
  year={2025}
}

@inproceedings{tang2025seeing,
  title={Seeing far and clearly: Mitigating hallucinations in mllms with attention causal decoding},
  author={Tang, Feilong and Liu, Chengzhi and Xu, Zhongxing and Hu, Ming and Huang, Zile and Xue, Haochen and Chen, Ziyang and Peng, Zelin and Yang, Zhiwei and Zhou, Sijin and others},
  booktitle={Proceedings of the Computer Vision and Pattern Recognition Conference},
  pages={26147--26159},
  year={2025}
}
}

\end{document}